\documentclass{article}

\usepackage{arxiv}

\usepackage[utf8]{inputenc} 
\usepackage[T1]{fontenc}    
\usepackage{hyperref}       
\usepackage{url}            
\usepackage{booktabs}       
\usepackage{amsfonts}       
\usepackage{nicefrac}       
\usepackage{microtype}      
\usepackage{lipsum}
\usepackage{graphicx}
\graphicspath{ {./images/} }

\usepackage{booktabs}
\usepackage{tabularx}
\usepackage{placeins}
\usepackage{makecell}
\usepackage{bm}
\usepackage{caption}

\usepackage{amsmath}  
\usepackage{graphicx} 
\usepackage{float}

\usepackage{makecell}

\usepackage{algorithm}
\usepackage{algpseudocode}
\usepackage{amsthm}
\usepackage{amssymb}
\usepackage{url}

\theoremstyle{definition}
\newtheorem{definition}{Definition}[section]
\usepackage{fix-cm}

\usepackage{ragged2e}

\title{Combining psychoanalysis and computer science: an empirical study of the relationship between emotions and the Lacanian discourses}

\author{
 Minas Gadalla \\
  Department of Computer Engineering \& Informatics\\
  University of Patras\\
  Patras, Greece\\
  \texttt{gkantalla@ceid.upatras,gr} \\
   \And
 Sotiris Nikoletseas \\
  Department of Computer Engineering \& Informatics\\
  University of Patras \& Computer Technology Institute and Press “Diophantus” (CTI)\\
  Patras, Greece \\
  \texttt{nikole@cti.gr} \\
  \And
 José Roberto de A. Amazonas \\
  Department of Computer Architecture\\
  Technical University of Catalonia - UPC\\
  Catalonia, Spain \\
  \texttt{jose.roberto.amazonas@upc.edu} \\
}

\begin{document}
\maketitle
\begin{abstract}
This research explores the interdisciplinary interaction between psychoanalysis and computer science, suggesting a mutually beneficial exchange. Indeed, psychoanalytic concepts can enrich technological applications involving the human factor, such as social media and other interactive digital platforms. By providing deeper insights into elusive, unconscious aspects of communication, psychoanalytic methods can enhance content-centric applications like fake news detection and mental health diagnostics. Conversely, computer science, especially Artificial Intelligence (AI), can contribute quantitative concepts and methods to psychoanalysis, identifying patterns and emotional cues in human expression. 


In particular, this research aims to apply computer science methods to establish fundamental relationships between emotions and Lacanian discourses. Such relations are discovered in our approach via empirical investigation and statistical analysis, and are eventually validated in a theoretical (psychoanalytic) way. It is worth noting that, although emotions have been sporadically studied in Lacanian theory, to the best of our knowledge a systematic, detailed investigation of their role is missing. Such fine-grained understanding of the role of emotions can also make the identification of Lacanian discourses more effective and easy in practise. In particular, our methods indicate the emotions with highest differentiation power in terms of corresponding discourses; conversely, we identify for each discourse the most characteristic emotions it admits. As a matter of fact, we develop a method which we call Lacanian Discourse Discovery (LDD), that simplifies (via systematizing) the identification of Lacanian discourses in texts.

Although the main contribution of this paper is inherently theoretical (psychoanalytic), it can also facilitate major practical applications in the realm of interactive digital systems. Indeed, our approach can be automated through Artificial Intelligence methods that effectively identify emotions (and corresponding discourses) in texts. 
\end{abstract}

\keywords{Psychoanalysis \and Computer Science \and Natural Language Processing (NLP) \and Lacanian Discourses \and Lacanian Discourse Discovery (LDD) \and Lacanian Discourse Analysis (LDA) \and Emotions}

\section{Introduction}
This research aims to further explore the potential interactions between Psychoanalysis and Computer Science, envisioning a cross-fertilization that can be mutually beneficial for both fields. This exchange is expected to be advantageous in both ``directions'',  from Psychoanalysis to Computer Science and vice versa. 

{\bf From Psychoanalysis to Computer Science:} we claim that psychoanalytic concepts and approaches can be very instrumental in Computer Science applications that heavily involve the human factor, such as in social media and other interactive digital platforms, systems, and tools. Indeed, in such text-based systems, psychoanalytic-based approaches have the potential to extract valuable, usually elusive information by providing insights into the underlying dynamics, motivations, and meanings of texts. They go beyond surface-level analysis and delve into the unconscious, possibly hidden aspects of communication. They can be applied to various types of texts and communication situations, providing a deeper understanding of the message being conveyed and the mechanisms that shape the content. Such insight can significantly benefit the effectiveness and performance of diverse digital applications and systems that are inherently content-centric; such systems include software tools for the detection of fake news in digital platforms, digital health applications for the early diagnosis of mental conditions and diseases, as well as software tools for deciding whether a given text is written by Artificial Intelligence (AI) programs or by humans.

{\bf From Computer Science to Psychoanalysis:} Sigmund Freud envisioned the future inclusion of quantitative, interdisciplinary concepts and methods (such as from Physics) into psychoanalysis. In a similar spirit, Jacques Lacan expressed and visualized psychoanalytic concepts using analogues from combinatorial mathematics (sets and networks) and mathematical logic (logical formulae); in particular, he thought of the language of the unconscious as a chain of quasi-mathematical inscriptions, similarly to a computer language as well as a cryptographic code. In this line, it is reasonable to assume that particularly Computer Science methods can be quite helpful in establishing solid quantitative contributions to Psychoanalysis. Particularly the use of AI is very relevant, in view of its ability, when properly used, to identify complicated or hidden patterns, emotional cues, and underlying themes in human speech or writing.

Regarding the first direction mentioned above, we have already conducted promising research, for the automated detection of fake news. In \cite{phycho_driven} we investigated the incorporation of the human factor and user perception in text-based interactive media, specifically focusing on the reliability of user texts compromised by behavioral and emotional dimensions. In particular, we designed a Psychoanalysis-based approach, using the notion of Lacanian discourse types to capture and understand the underlying characteristics and qualities of texts (news headlines) and how they inherently relate to real or fake news type of texts. The approach involves first identifying Lacanian discourses in a text (news headlines), and then using an algorithm for predicting whether this text is real or fake depending on the type of Lacanian discourses present in it. The performance evaluation demonstrated quite high effectiveness and accuracy of this Psychoanalysis-based prediction compared to standard methods. As far as we know, this was the first time computational methods were systematically combined with Psychoanalysis.

Following up this significant first step, the research work discussed here addresses the ``inverse direction'', of applying Computer Science methodologies to Psychoanalysis. In particular, the purpose is to empirically investigate (and then theoretically validate) possible fundamental relations between emotions and Lacanian discourses. J. Lacan himself highlighted the significance of a few fundamental emotions (such as anxiety and anguish, \cite{soler2016lacanian}). Recent research \cite{bucci2022} has investigated the important role of some emotions (affects) in Lacanian theory \cite{lacan19681969seminaire}, \cite{lacan1974seminaire}. However, to the best of our knowledge, there is still no work in the state of the art that systematically relates emotions with corresponding Lacanian discourses. This work aims to address this gap, via empirically (statistically) investigating this relation in a systematic, fine-grained manner. Such solid understanding of the relation between emotions and Lacanian discourses is important because it will make the identification of Lacanian discourses in texts easier and more robust, since it will be based on the (easier to grasp) presence of emotions in texts.

The method basically includes the following key steps: a) the emotions in a given text are identified, from a well-known, fine-grained set of 30 emotions; b) the Lacanian discourses in the text are also identified; c) a statistical investigation is performed to identify the potentially inherent relation between emotions and Lacanian discourses: for each discourse, which are the most characteristic emotions it admits? Which emotions exhibit the highest differentiation power in terms of corresponding discourses? d) finally, these statistical findings are theoretically (psychoanalytically) validated.

The main contribution of this research is a psychoanalytic one per se: the establishment of a systematic relation among emotions and Lacanian discourses, the introduction of the concept of Lacanian Discourse Discovery (LDD) which systematizes the discovery of the Lacanian discourses. It is worth noting that this method can be automatized to a great extent, since current computer-based methods (primarily employing AI systems and tools) can be used to effectively detect emotions in texts. This way, there is great potential for developing effective, real-world applications, based on the automated identification of emotions and corresponding discourses.

After this Introduction, in Section \ref{sec:theoretical} is presented a theoretical framework that includes how emotions appear in S. Freud's and J. Lacan's works, a review of the five Lacanian discourses a description of the Lacanian Discourse Analysis (LDA) and a review of the classification of emotions as proposed by different authors. In Section \ref{sec:relatedwork} a brief presentation of related works is provided. In Section \ref{sec:adopted_methodology} the adopted methodology is presented in detail. The results and the corresponding discussion are presented in Section \ref{sec:results}. Section \ref{sec:discussion} concludes the paper and indicates future directions.

\section{Theoretical Framework}
\label{sec:theoretical}
The objectives of this section are to review how emotions appear in S. Freud's and J. Lacan's works, to review the five Lacanian discourse, to describe the Lacanian Discourse Analysis (LDA) and to review the major proposed classification of emotions. 

\subsection{Emotions in Freud’s works}
\label{sec:emotions_freud}
According to \cite[pp 9-10]{barrett2017emotions} (cited \emph{verbatim}):
\begin{quote}
\textit{The classical view of emotion holds that we have many such emotion circuits in our brains, and each is said to cause a distinct set of changes, that is, a fingerprint. Perhaps an annoying coworker triggers your ``anger neurons'', so your blood pressure rises; you scowl, yell, and feel the heat of fury. Or an alarming news story triggers your ``fear neurons'', so your heart races; you freeze and feel a flash of dread. Because we experience anger, happiness, surprise, and other emotions as clear and identifiable states of being, it seems reasonable to assume that each emotion has a defining underlying pattern in the brain and body.   } 
\end{quote}

Still in \cite[pp 113-114]{barrett2017emotions} (cited \emph{verbatim}):
\begin{quote}
\textit{Affect is the general sense of feeling that you experience throughout each day. It is not emotion but a much simpler feeling with two features. The first is how pleasant or unpleasant you feel, which scientists call valence. The pleasantness of the sun on your skin, the deliciousness of your favorite food, and the discomfort of a stomachache or a pinch are all examples of affective valence. The second feature of affect is how calm or agitated you feel, which is called arousal. The energized feeling of anticipating good news, the jittery feeling after drinking too much coffee, the fatigue after a long run, and the weariness from lack of sleep are examples of high and low arousal. Anytime you have an intuition that an investment is risky or profitable, or a gut feeling that someone is trustworthy or an asshole, that’s also affect. Even a completely neutral feeling is affect.
}
\end{quote}

In the literature in general, and in the psychoanalytic literature in particular, many times both terms are used as synonyms of each other, or one is mistakenly used by the other.

It is not a purpose of this work to delve into the intricacies of each of these terms. This paper deals mainly with emotions. When the term affect appears, it means that this word was the original choice of the referenced author.

It is outside the scope of this work to present a comprehensive review of the Psychoanalysis theory. In fact, it is assumed that the reader is at least acquainted with its fundamental concepts as the Unconscious, the triad (ego, id, superego) and the theory of sexuality, for example. 

Early in 1893 - 1895, J. Breuer and S. Freud \cite{freud1895}, in their work about Hysteria, stated that:
\begin{quote}
 \textit{Any experience which calls up distressing \textbf{affects} - such as those of fright, anxiety, shame or physical pain - may operate as a trauma of this kind;}   
\end{quote}

Further on, in the same publication:
\begin{quote}
    \textit{For we found, to our great surprise at first, that each individual hysterical symptom immediately and permanently disappeared when we had succeeded in bringing clearly to light the memory of the event by which it was provoked and in arousing its accompanying \textbf{affect}, and when the patient had described that event in the greatest possible detail and had \textbf{put the affect into words}.}
\end{quote}

About the same time, in 1894, S. Freud published ``The Neuropsychoses of Defence'' \cite{freud1894} in which he exposed the theory that the neuropsychoses arise from an experience, idea or feeling that produces a powerful distressing emotion as reproduced below:
\begin{quote}
   \textit{For these patients whom I analysed had enjoyed good mental health up to the moment at which an occurrence of incompatibility [\underline{Unverträglichkeit]} took place in their ideational life – that is to say, until their ego [\underline{Ich}] was faced with an experience, an idea or a feeling which aroused such a \textbf{distressing affect} that the subject [Person] decided to forget about it because he had no confidence in his power to resolve the contradiction between that incompatible idea and his ego by means of thought activity.} 
\end{quote}

In face of such incompatible ideas the patients try to ``push the thing away'', make an effort of not thinking of it, of suppressing it. When this kind of ``forgetting'' did not succeed it led to various pathological reactions which produced either hysteria or an obsession or a hallucinatory psychosis. The memory trace and the \textbf{affect} which is attached to the idea are there once and for all and cannot be eradicated. 

According to S. Freud:
\begin{quote}
    \textit{Up to this point the processes in hysteria, and in phobias and obsessions are the same; from now on their paths diverge. In hysteria, the incompatible idea is rendered innocuous by its sum of excitation being transformed into something somatic.}

    \textit{If someone with a disposition [to neurosis] lacks the aptitude for conversion, but if, nevertheless, in order to fend off an incompatible idea, he sets about separating it from its \textbf{affect}, then that \textbf{affect} is obliged to remain in the psychical sphere. The idea, now weakened, is still left in consciousness, separated from all association. But its \textbf{affect}, which has become free, attaches itself to other ideas which are not in themselves incompatible; and, thanks to this ``false connection'', those ideas turn into obsessional ideas.}
\end{quote}
 
In 1915, S. Freud published his fundamental work about the Unconscious \cite{freud1915e} in which he postulates the destination of the repressed \textbf{affect}:
\begin{quote}
    \textit{The importance of the system Cs. (Pcs.) as regards access to the release of affect and to action enables us also to understand the part played by substitutive ideas in determining the form taken by illness. It is possible for the development of \textbf{affect} to proceed directly from the system Ucs.; in that case the \textbf{affect} always has the character of \underline{anxiety} \textbf{[anguish]}, for which all ``repressed'' \textbf{affects} are exchanged.} 
\end{quote}
\noindent{in which Cs., Pcs. and Ucs stand for the Conscious, Preconscious and Unconscious systems respectively.}

The original version of the aforementioned paragraph, in German, is the following:
\begin{quote}
    \textit{Die Bedeutung des Systems Bw (Vbw) für die Zugänge zur Affektentbindung und Aktion macht uns auch die Rolle verständlich, welche in der Krankheitsgestaltung der Ersatzvorstellung zufällt. Es ist möglich, daß die Affektentwicklung direkt vom System Ubw ausgeht, in diesem Falle hat sie immer den Charakter der \underline{Angst}, gegen welche alle ``verdrängten'' \textbf{Affekte} eingetauscht werden.}
\end{quote}
\noindent{where it can be seen that the word used by S. Freud to designate the transformed repressed emotion is \underline{Angst}.}

The Spanish version of the Freud's complete works \cite{freud1915spanish}, translated by Luis López Ballesteros y de Torres and formally approved by S. Freud brings the following paragraph:
\begin{quote}
    \textit{La significación del sistema Cc. (Prec.) con respecto al desarrollo de afecto y a la acción, nos descubre la de la representación sustitutiva en la formación de la enfermedad. El desarrollo de afecto puede emanar directamente del sistema Inc., y en este caso, tendrá siempre el carácter de \underline{angustia}, la cual es la sustitución regular de los afectos reprimidos.}
\end{quote}
\noindent{where it can be seen that \emph{Angst} has been translated by \emph{angustia}, i.e., \emph{anguish} in English.}

It is important to take into account the Editor's Appendix: The Term ``Angst'' and its English Translation \cite{freud1915editorappendix} that states the following:
\begin{quote}
\textit{    The word universally, and perhaps unfortunately, adopted for the purpose has been ``anxiety'' – \textbf{unfortunately}, since ``anxiety'' too has a current everyday meaning, and one which has only a rather remote connection with any of the uses of the German ``\underline{Angst}''.}
    \begin{center}
        
    .
    
    .
    
    .
    \end{center}
    
\textit{    A still more acute condition is described in English by the word ``anguish'', which has the same derivation; and it is to be remarked that Freud in his French papers uses the kindred word ``angoisse'' (as well as the synonymous ``anxiété'') to render the German ``Angst''.}
\end{quote}

In this work we stay with the translation proposed by the Spanish version and will refer to \textbf{anguish} as the result of the repressed \textbf{affect}.

The topic of \textbf{anguish} [``Angst''] has been a permanent concern of S. Freud throughout his life and it has appeared in most of his writings. It is out of the scope of this work to delve deeper into this theme.

In summary, to conclude, it may be said that:
\begin{itemize}
    \item S. Freud uses the term \textbf{affect} to collectively refer to emotions;
    \item Emotions are essential elements of the aetiology of neuropsychoses;
    \item If the emotions were repressed or not along the distressing experience determines the development path of the neuropsychosis;
    \item No single emotion is identified as more important in the development of neuropsychoses;
    \item All repressed and not discharged emotions are exchanged by \textbf{anguish}.
\end{itemize}

\subsection{Emotions in Lacan's works}
\label{sec:emotions_lacan}
In spite of their importance,  S. Freud did not consider emotions appropriate to the work of interpretation and a path to access the unconscious. The representations (\emph{Vorstellungen}) and the representatives of representations (\emph{Vorstellungsrepräsentanzen}) are the elements truly repressed in the unconscious that must again be discovered by means of deciphering.  The \emph{Vorstellungsrepräsentanzen} are strictly equivalent to J. Lacan's signifiers. Affects are situated along the pleasure - unpleasure axis, are not completely repressed having become disconnected from the original trauma and being able to move around different \emph{Vorstellungsrepräsentanzen}. In this way, affects sliding from one representation to another lie (according to J. Lacan \cite{lacan1958seminaire}, class given on November 26, 1958) about their origins. In other words, affects drift away from their original anchoring point in a traumatic sexual experience that was very real. According to C. Soler in \cite{soler2016lacanian}:
\begin{quote}
    \textit{Placing at the beginning of mankind's fate the experience of an unmasterable excitation that overwhelms the subject and generates anguish that he qualifies as ``real'', he bestows a very specific status on \textbf{anguish}: it is both effect and cause. It is the effect of a real encounter with the said excitation, but it is the cause of the repression that will generate symptoms and resonate in subsequent affects, first among which is ``\textbf{anguish} as a signal'', which is both a memorial and a warning: a memorial of the first trauma and a warning about an imminent danger.}
\end{quote}

However, in disagreement with S. Freud, J. Lacan stated that affects cannot be conceived outside the Symbolic realm and therefore are in the field of the psychoanalytic technique. In fact, J. Lacan devoted one whole year seminar \cite{lacan2004seminaire} to \textbf{anguish} to demystify it as the end of analysis.

To start with, J. Lacan in \cite{lacan2004seminaire} posited that \textbf{anguish} is the affect that does not lie or does not deceive (\textit{ne trompe pas}). It involves major bodily sensations, as that one has a stone in one's throat or that one’s heart is running too fast. \textbf{Anguish} has three characteristics:
\begin{enumerate}
    \item there is a blurred threat;
    \item it is experienced;
    \item the subject knows that it concerns him/her but has no explanation for it.
\end{enumerate}

\textbf{Anguish} does not drift around signifiers, it remains attached to the original cause. It is not without an object, its object is the object \textit{a} (\emph{le petit a}) and it indicates the oncoming arrival of something that is real. As it is tied to something that is real, it becomes an ally to interpretation.

According to J. Lacan, as exposed in \cite{soler2016lacanian}:
\begin{quote}
    
\begin{itemize}
    \item St. Thomas Aquinas mentions fear, not \textbf{anguish}, in Summa Theologica. Christian theology situates emotions in the relationship between a man and the God's divine will, the articles of faith saying precisely what He wants, it is understandable why there would be more fear than \textbf{anguish}. 

    \item In 1512 Luther, in his doctrine of the man inhabited by the radical evil that even Christ is not able to erase, salvation is still possible, a salvation that is only granted by the arbitrary will of God, from whom one no longer knows what to expect. It is the ``birth'' of \textbf{anguish}.

    \item Pascal was a mathematician, a scientist, an inventor and a mystic. He revived the idea of the absolute rule of the will of the divine God (the Other) and that of the defenselessness and absolute impotence of man who can do nothing for his own salvation. Some will be saved, others will not be; it is already written in God's obscure plans about which we know not what inspires them. 

    \item Pascal formulates a newly generated \textbf{anguish}, ``\textit{Le silence éternel de ces espaces infinis m’effraie}'' (``The eternal silence of these infinite spaces terrifies me''). This sentence indicates that the fear of the thundering voice (which was terrible but ever so reassuring owing to its presence) had already been replaced by doubt, highlighting the silence of a voice that was no longer heard.

    \item Capitalism has replaced symbolic productions by the objects it produces. People talk a great deal about the rise of depression in our era, but the true mood illness of capitalism is \textbf{anguish}. \textbf{Anguish} is the emotion tied to ``subjective destitution''; it is an affect that arises when the subject perceives himself as an object. Scientific capitalism, with its technological effects, brings about destitution far more radically than psychoanalysis does: it uses and abuses subjects as instruments. 
\end{itemize}

\end{quote}

The presentation about \textbf{anguish} could go on and on but it should by now be clear that it is the fundamental emotion for psychoanalytic considerations.

J. Lacan went far beyond the consideration of \textbf{anguish}, by developing a complete ``Theory of affects'' in which he states that there are no known affects that are lacking in a bodily component. Thus in order to conceptualize affect, one must ``include the body'' \cite{lacan1974television} and as exposed in \cite{soler2016lacanian}:

\begin{quote}
The organic individual that serves as a support for the speaking subject represented by the signifier is not the body. There are:
\begin{enumerate}
    \item the living organism, which is the object studied by biology and which psychoanalysts need know little about; 
    \item the subject defined by his speech; and 
    \item the body the subject has that is also studied by psychoanalysis since it is subject to symptoms.
\end{enumerate}
\end{quote}

C. Soler goes on \cite{soler2016lacanian}:
\begin{quote}
The conversion phenomena that Freud brought to light thanks to hysteria must be generalized: the body is corporized in a ``signifying way'' (Lacan, 1998a, p. 26) and the jouissances of speaking beings are jouissances that are converted into language, they are affected by the ciphering that goes on in the unconscious, the affected party being the bodily individual in his flesh. J. Lacan posits that the signifier affects something other than itself: it affects the bodily individual that is thereby made into a subject. The subjectifying effect of language, which was very quickly highlighted as involving a loss, is tied to other effects in the real that serve as regulators of the jouissance that symptoms involve and that have consequences as regards the sexual relationship.

The first affecting party is thus language, and the affected party is not simply the imaginary body, but its capacity to enjoy, jouissance being the only substance that psychoanalysis deals with. 

In addition to the effect of language there are the collective effects of what Freud called ``civilization,'' which Lacan renamed ``discourse'' in order to indicate that the structure of language is no less inscribed in our social reality than in the unconscious, that it determines social bonds there, and that it presides in each era over the economy of bodies, over the regulating of interpersonal relations, and consequently over the configuration of the dominant affects in a given era.
\end{quote}

J. Lacan presented in Television an affect series that is both unique and surprising \cite{lacan1974television}. 

According to C. Soler \cite{soler2016lacanian}:
\begin{quote}
This series does not seek to cover all affects, but specifically those that are responses to the reality of the unconscious (\textit{au réel de l’inconscient}) – in other words, to the impossible relationship between the sexes – and to its effects, which psychoanalysis alone sheds light on. 

\begin{center}
        
    .
    
    .
    
    .
    \end{center}
    
From this vantage point, not all affects are equivalent. Of the four I have discussed, the first two (\textbf{sadness} and \textbf{joyful knowledge}) are related to knowledge, and the last two (\textbf{boredom} and \textbf{moroseness}) are related to sex. The series is thus organized, on the one hand, as a function of one's ethical position in relation to knowledge – this is the difference between sadness and joyful knowledge – and on the other hand, as a function of the historicity of discourse, with ``our'' boredom and ``our'' moroseness, the typical dominance of the latter corresponding to the effects of the unconscious on the body when the reparative semblances that nourished Eros are missing.
\end{quote}

In conclusion, it seems then reasonable try to identify if there are recognizable relationships between emotions and the different discourses that govern the social bonds.

\subsection{The Lacanian discourses}
\label{sec:lacanian_discourses}
\makeatletter
\newcommand{\superimpose}[2]{%
  {\ooalign{$#1\@firstoftwo#2$\cr\hfil$#1\@secondoftwo#2$\hfil\cr}}}
\makeatother

\newcommand{\veewedge}{\mathpalette\superimpose{{\vee}{\wedge}}}
\newcommand{\lessgreater}{\mathpalette\superimpose{{<}{>}}}
\newcommand{\strikeQ}{\mathpalette\superimpose{{\text{---}}{Q}}}

\newcommand{\Xarrows}{\mathpalette\superimpose{{\nearrow}{\nwarrow}}}

\newcommand{\matheme}[4]{\left\uparrow\dfrac{#1}{#2} \hspace{.3cm}\overrightarrow{\Xarrows}\hspace{.3cm} \dfrac{#3}{#4}\right\downarrow}

\newcommand{\master}{\matheme{\text{S1}}{\$}{\text{S2}}{a}}

\newcommand{\university}{\matheme{\text{S2}}{\text{S1}}{a}{\$}}

\newcommand{\analyst}{\matheme{a}{\text{S2}}{\$}{\text{S1}}}

\newcommand{\hysteric}{\matheme{\$}{a}{\text{S1}}{\text{S2}}}

\newcommand{\discourses}{\matheme{\text{Agent}}{\text{Truth}}{\text{other}}{\text{Production}}}

\newcommand{\newdiscourses}{\matheme{\text{Semblance}}{\text{Truth}}{\text{Jouissance}}{\text{Surplus-jouissance}}}

\newcommand{\newmatheme}[4]{\left\downarrow\dfrac{#1}{#2} \hspace{.3cm}\Xarrows\hspace{.3cm} \dfrac{#3}{#4}\right\downarrow}

\newcommand{\capitalist}{\newmatheme{\text{\$}}{\text{S1}}{\text{S2}}{{a}}}

\newcommand{\nonrapport}[4]{\dfrac{#1}{#2} \hspace{.3cm}\overrightarrow{\blacktriangle}\hspace{.3cm} \dfrac{#3}{#4}}

\newcommand{\nonrapdiscourses}{\nonrapport{\text{Agent}}{\text{Truth}}{\text{other}}{\text{Production}}}

\newcommand{\nonrapnewdiscourses}{\nonrapport{\text{Semblance}}{\text{Truth}}{\text{Jouissance}}{\text{Surplus-jouissance}}}

J. Lacan proposed his four discourses in 1969 in his seminar L’envers de la psychanalyse \cite{lacan1969seminaire}. He also developed his ideas further in following seminars as \cite{lacan1972seminaire} and \cite{lacan1974seminaire}. His ideas were extremely well explained by L. Bailey in \cite{bailly}.

The Four Discourses theory constitutes a formalism of the different ways people relate to each other and the economy of knowledge and enjoyment in social relationships. The general structure of is represented by a \textit{matheme} as shown in Eq. (\ref{eq:discourses}).

\begin{equation}
    \label{eq:discourses}
    \discourses
\end{equation}

The transmitter of the discourse is the \textit{Agent} that occupies the top left position of the \textit{matheme}. The \textit{Agent} addresses the \textit{other} that occupies the top right position of the \textit{matheme}. The \textit{other} is the receiver of the message. This interaction is represented by the arrow pointing from the \textit{Agent} to the \textit{other}. What is said by the \textit{Agent} is driven by the \textit{Truth} that is hidden below the \textit{Agent}, belongs to the Unconscious and is not directly accessible. This driving is represented by the upward arrow that points from the \textit{Truth} to the \textit{Agent}. The hidden \textit{Truth} is embedded in the transmitted message. This is represented by the oblique arrow that points from the \textit{Truth} to the \textit{other}. The message is interpreted, consciously and unconsciously by the \textit{other} who will deliver a response named the \textit{Production}. The \textit{Production} is the fourth element of the \textit{matheme} that occupies the bottom right position of the \textit{matheme}. The  \textit{Production} is delivered to the \textit{Agent} and this is represented by the oblique arrow that points from the \textit{Production} to the \textit{Agent}.

There isn't any arrow pointing from the \textit{Production} to the \textit{Truth} indicating that the provided response is incomplete, unsatisfactory. It is impossible to establish a perfect communication between the transmitter and receiver sides. This impossibility is caused by the fact that people are speaking beings and the communication is limited by the constraints imposed by the language. Nevertheless, this impossibility is what creates and maintains the social bonds between beings.

The four elements of the \textit{matheme} are placeholders for the roles assumed by the interacting parties in the communication process, namely:

\begin{itemize}
 	\item {\bf S1}: the master signifier represents the true essence of the subject, it organizes both the psychical and everyday reality. It may be summarized by \emph{who I am}.
	\item {\bf S2}: represents the knowledge of the subject. It may be summarized by \emph{what I know}.
	\item {\bf a}: represents the object cause of desire. It may be summarized by \emph{what I want}.
	\item {\bf \$}: represents the barred subject castrated by the language. It may be summarized by \emph{what I speak}.
 \end{itemize}

The four discourses are defined by the position each element occupies in the general \textit{matheme} representation shown in Eq. (\ref{eq:discourses}).

\vspace{2mm}

$\master$ {\bf Master Discourse:} the \textit{Agent} position is occupied by {\bf S1} who addresses the \textit{other} not as a person \textit{per se} but as the holder of a knowledge ({\bf S2}). The message may convey not only traces of authority and power but also the image the \textit{Agent} builds about him/herself to be publicly disclosed. {\bf S1} is driven by {\bf \$} meaning that the message suffers from the limitations imposed by the language. The receiver {\bf S2} cannot fully enjoy the possession of knowledge that is delivered to the \textit{Agent} as the \textit{Production} represented by {\bf a}.

A counterclockwise rotation of the elements of the {\bf Master Discourse} \textit{matheme} leads to the second discourse.

\vspace{2mm}

$\university$ {\bf University Discourse:} in this discourse {\bf S1} is downgraded and becomes the hidden \textit{Truth}. {\bf S2} becomes the \textit{Agent}. The transmitted message may convey verified and referenced information. The \textit{Agent} may be a real expert or authority in his/her field of work or can be just the false image of an expert whose expertise is derived by the supporting {\bf S1}. It is a discourse that represents institutions and their manipulation actions upon the receivers of the message. The \textit{Agent} addresses {\bf a}, i. e., addresses the receiver's desire to be accepted by the prestigious {\bf S1} and to incorporate some of its characteristics. The end result, i. e., the \textit{Production}, however, is just more submission of the receiver to the transmitter.

A counterclockwise rotation of the elements of the {\bf University Discourse} \textit{matheme} leads to the third discourse.

\vspace{2mm}

$\analyst$ {\bf Analyst Discourse:} in the discourse of the Analyst, the \textit{Agent} becomes the {\bf a} of the other, meaning that he/she becomes the other's desire to know more about him/herself. The \textit{Agent} becomes the subject-supposed-to-know ({\bf S2}) the \textit{Truth} about the other. The \textit{Agent} addresses the other as a speaking being, i.e., employing a neutral curiosity, free of judgement aiming at provoking the other to generate a \textit{Production} that reveals his/her master signifiers {\bf S1}. This chain of signifiers may be interpreted and are reflected back, as if by an empty mirror, to the other that ends up knowing more about him/herself. In a long term relationship the other ends up learning that all knowledge acquired came from him/herself and that the \textit{Agent} is no more necessary.

Another counterclockwise rotation of the elements of the {\bf Analyst Discourse} \textit{matheme} leads to the final fourth discourse.

\vspace{2mm}

$\hysteric$ {\bf Hysteric Discourse:} one doesn't have to be hysterical in the clinical sense to hold the discourse of the Hysteric; indeed, J. Lacan made it clear that this type of discourse in non-hysterical people, is precisely what leads to true learning. The \textit{Agent} as a speaking being {\bf \$}, driven by his/her desire to know {\bf a},   addresses the \textit{other} in his social role {\bf S1} as someone that is able to produce true information and knowledge {\bf S2}, the targeted \textit{Production}. The \textit{Production} is usually unsatisfactory making the \textit{Agent} to continue the questioning till the knowledge {\bf S2} of the \textit{other} is exhausted.

In Seminar XVIII \cite{lacan1971seminaire} and Seminar XIX \cite{lacan19711972seminaire} J. Lacan reformulated the general structure of the discourse's \textit{matheme} that became as shown in Eq. (\ref{eq:newdiscourses}).

\begin{equation}
    \label{eq:newdiscourses}
    \newdiscourses
\end{equation}

The \textit{Agent} becomes the \textit{Semblance} to indicate that the discourse is defined by the position or role taken by someone in relation to the \textit{other}. For example, the discourse of the {\bf Master} takes shape if someone plays the role of the commanding agent. The position of the \textit{other} is substituted by the \textit{Jouissance} that is defined by J. Lacan in Seminar XIX \cite{lacan19711972seminaire} as a disturbing dimension in the experience of the body. The subject is unable to experience itself as a self-sufficient enjoying entity. The enjoyment is conditioned by the addressing \textit{Semblance} which is expected to manage it.

Last but not least, the \textit{Production} of the discourse becomes the \textit{Surplus-jouissance}. J. Lacan borrows the Marx's concept of ``surplus value'' \cite{marx2018capital} to build the concept of ``surplus-jouissance''. ``Surplus value'' is defined as the difference between the exchange value of products of labor (commodities) and the value that coincides with the effort of producing these products, i.e., the means of production and labor power. Within the capitalist system gaining surplus value is the sole objective, profit making and expansion of the capital are the driving motors.

J. Lacan states that the general structure of a discourse is homologous to the system of capitalism described by Marx. While in the capitalist production the surplus values and/or commodities are fetishized, in the use of discourse a fetishist relation with surplus-jouissance is created \cite{lacan19681969seminaire}.

In the use of discourse, language is produced. The attempt to address jouissance by means of language produces an extra of corporeal tension that is beyond the language itself. Such surplus-jouissance can only be located in the realm of fantasy or delusion.

By the end of the years 1960, J. Lacan started to comment about a fifth discourse, the \textbf{Capitalist Discourse}, highlighting the differences from the other four discourses. In 1972, in a lecture given the University of Milan \cite{contri2008}, J. Lacan presented the precise structure of the \textbf{Capitalist Discourse} that is shown in Eq. (\ref{eq:capitalist}).

\begin{equation}
    \label{eq:capitalist}
    \capitalist
\end{equation}

At first sight, the \textbf{Capitalist Discourse} appears to be a simple variation of the \textbf{Master Discourse} but, in fact, it is a disruptive mutation of it. Comparing both structures it is possible to identify three differences:
\begin{enumerate}
    \item {\bf \$} and {\bf S1} exchange places.
    \item The arrow pointing upward on the left that makes the position of the truth inaccessible  in the classic discourse changes into an arrow pointing downwards.
    \item The horizontal arrow that established the connection between the \textit{Agent} and the \textit{other}, or between the \textit{Semblance} and the \textit{Jouissance} disappears.
\end{enumerate}

In the \textbf{Master Discourse}, {\bf S1} organizes the discourse due to its position as the \textit{Agent} (\textit{Semblance}). In the \textbf{Capitalist Discourse}, {\bf S1} has been degraded to occupy the \textit{Truth} position that is not hidden anymore but has become accessible as indicated by the arrow pointing downwards on the transmitter's side. In the \textbf{Capitalist Discourse} the castrated subject {\bf \$} does not express its needs or demands by addressing the \textit{other}. {\bf \$}, by itself, elects a fetishized commodity to fulfill the needs or demands. The materialization of the fetish happens by the acquisition of an asset ({\bf S2}) that most certainly does not match the subject's expectation. This frustration fuels the \textit{Production} and the object-cause-of-desire {\bf a} feedbacks to  {\bf \$}. The subject will try to combat the frustration electing a new {\bf S1} and the cycle repeats itself.

The cycle {\bf \$} $\rightarrow$ {\bf S1} $\rightarrow$ {\bf S2} $\rightarrow$ {\bf a} $\rightarrow$ {\bf \$} $\rightarrow$ $\dots$ can continue indefinitely. This a cycle that represents serious disorders such as addiction and compulsion. Addiction includes drugs, social media, pornography and others. Compulsion includes buying, working, sex and others.

The attachment to the \textbf{Capitalist Discourse} is stimulated by the big corporations, media and marketing giants using all kinds of manipulation tools to enslave people to the carousel of consumerism. The Subject is reduced to a mere object to be used and discarded.

In the four classic discourses the social bonds are sustained by the impossibility of communication between the hidden \textit{Truth} and the \textit{Production}. In J. Lacan terms, the non-rapport ($\blacktriangle$) is at the basis of the relationship between the \textit{Agent} (\textit{Semblance}) and the \textit{other} (\textit{Jouissance}). 

\[\nonrapdiscourses\]

\[\nonrapnewdiscourses\]

In the \textbf{Capitalist Discourse} the social bonds are destroyed.

\subsection{Lacanian Discourses Analysis (LDA) and Lacanian Discourses Discovery (LDD)}
\label{sec:lda}
Lacanian Discourse Analysis (LDA) is not a system or methodology developed by Lacan himself but rather a way of utilizing Lacanian concepts to analyze texts. It is considered a subsequent attempt by scholars, especially in the psychosocial domain, to create a ``system'' of text analysis aligned with Lacanian psychoanalytic theory, which was originally designed for therapeutic sessions. However, this paper does not aim to delve deeply into LDA or engage with the intricacies of Psychoanalysis. Instead, it focuses on utilizing certain aspects of LDA, particularly the one of the five Lacanian Discourses, for a specific application and goal. To reflect this distinction, we refer to this approach as \textit{Lacanian Discourse Discovery} (LDD) rather than LDA.

The term ``Discovery'' is chosen deliberately to emphasize that this process of identifying the five discourses in text does not involve any analytical or methodological procedures, such as those outlined by Ian Parker in \cite{Parker2010}. Instead, the identification of the discourses is viewed as a discovery process, one that quickly and effectively uncovers the Lacanian discourses in text (in this work textual conversations)  without delving deeply into the attributes of the text or the characteristics of the speakers.

To the best of our knowledge, this work represents the first attempt to apply the Lacanian Discourses within the Natural Language Processing (NLP) domain of computer science; breaking down text or speech into smaller parts that computer programs can easily understand\footnote{NLP is not to be confused with Neuro-Linguistic Programming Therapy.} \cite{nlpintrod}. 


The approach of utilizing Lacanian aspects for text analysis is common in theoretical studies of philosophy and politics \cite{Wetherell1999}. For example, in \cite{Parker2010} the author  employs aspects of Lacan's work to provide a comprehensive analysis of an interview from a film, using psychoanalytic theory as a lens to explore and highlight connections and meanings within the text without imposing it as the definitive interpretation.

Structured text, like a movie script written by a screenwriter for production and audience entertainment, has the advantage of allowing one to map the social, cultural, and situational factors influencing the text and the relationships between the characters. In the same way, such an approach with less structured, richer texts is attempted. By no means should this be considered a new approach to psychoanalysis but rather a new, radical approach to combine Lacanian psychoanalytic concepts in the NLP domain. This approach aims to quickly identify the five Lacanian discourses when there is no information about social or cultural factors, the gender or attributes of the subjects, or when there are no clear connections or interactions between the characters. Additionally, this method is beneficial when the text is not structured to convey information, provide knowledge, or indirectly communicate with the reader. In this case, short everyday conversations between two speakers in text-only format have been used (see Section \ref{sec:work_only_text}).

By bringing parts of the Lacanian Discourses theories into the NLP domain, enabled to achieve:
\begin{enumerate}
    \item An efficient method for discovering Lacanian discourses in short, unstructured texts such as everyday dialogues.
    \item A foundation for the development of future computer-based applications using the Lacanian Discourses.
    \item For the first time, the integration of Lacanian discourse concepts with NLP tasks in the field of Computer Science.
\end{enumerate}

\subsection{Classification of emotions}
\label{sec:classification_of_emotions}
Human beings tend to think in categories, classes, and groups. It is necessary to categorize concepts and recurring states, as it helps to learn concepts, facilitates communication, makes useful predictions about our world, and creates mental building blocks for expressing more sophisticated thoughts \cite{goldstone_concepts}. The concept of categorization begins in infancy and is developed, expanded, and improved throughout one's lifespan \cite{norman_dual_2009}. Due to this inherent nature, categorization has been expanded to various scientific fields such as social sciences (e.g., identity formation and group behaviors \cite{hirschfeld_how_2004}), psychology, computer science (e.g., algorithms, data handling of computer systems and architectures), and linguistics (commonly known as features \cite{ekman_argument_1992}).

The categorization of emotions stands out as particularly complex and debated. Unlike more concrete categories (e.g., colors), emotions present unique challenges due to their subjective, multifaceted nature, and linguistic constraints. The study of emotions spans multiple disciplines, including psychology, neuroscience, and computer science, each bringing its perspective to understanding how emotions can be categorized and utilized.

Probably, the most basic set of emotions that could be found in human text is from a Chinese encyclopedia compiled in the first century B.C:
What are the feelings of men? They are joy, anger, sadness, disliking, and liking. These five feelings belong to men without their learning them (Chai \& Chai, 1885/1967, p. 379).

Such a basic scheme could potentially be valid and used as an early stage discovery of a state-of-the-art set of emotions to be used universally. Progressively, the most well-known schema are Ekman's \cite{ekman_selection_2008}, Plutchik's \cite{plutchik_psychoevolutionary_1980}, Circumplex theory of affect \cite{watson_consensual_1985}, EARL (HUMAINE \cite{humaine_2006}), and WordNet–Affect \cite{sedding_wordnet_2004}. According to \cite{arribas_ayllon_utility_2019}, Ekman's six basic emotions emerged as the most useful classification scheme for emotive language analysis in terms of ease of use by human annotators and training supervised machine learning algorithms. However, it has significant shortcomings, particularly in the representation of positive emotions. Plutchik's wheel of emotions \cite{plutchik_psychoevolutionary_1980} provides a rich emotional spectrum but is complex and has lower machine learning performance. The Circumplex model \cite{watson_consensual_1985} offers a dimensional representation of emotions, capturing nuances well but also presenting complexity and moderate machine learning performance. EARL \cite{humaine_2006}, designed for technological contexts, covers a wide range of emotions but has lower inter-annotator agreement and performance. Finally, WordNet–Affect \cite{sedding_wordnet_2004} is a rich lexical resource but is difficult to navigate and achieves the lowest agreement and performance.

Table \ref{tab:emotionsclassification} summarizes the advantages and disadvantages of different emotion classification schema.

\begin{table*}[t]
\centering
\setlength{\tabcolsep}{10pt} 
\caption{Summary of advantages and disadvantages of different emotion classification schema.}
\label{tab:emotionsclassification} 
\begin{tabularx}{\textwidth}{p{4cm}>{\RaggedRight}X >{\RaggedRight}X}
\toprule
\textbf{Scheme} & \textbf{Advantages} & \textbf{Disadvantages} \\
\midrule
Six Basic Emotions & Easy to understand and use; High agreement among annotators; Good overall performance. & Simplistic; It does not capture the full range of positive emotions. \\
\midrule
Wheel of Emotions & It provides a rich variety of emotions; Good agreement among annotators. & Complex; Lower performance in practical applications; Some emotions are debatable. \\
\midrule
Circumplex & It captures emotional nuances well; Reasonable agreement among annotators. & Complex; Moderate performance in practical applications; Overlapping and ambiguous categories. \\
\midrule
EARL & Designed for technological contexts; It covers a wide range of emotions. & Lower agreement among annotators; Lower performance in practical applications; Categories may not be distinct enough. \\
\midrule
WordNet--Affect & Rich in emotional vocabulary; Detailed emotional hierarchy. & Lowest agreement among annotators; Lowest performance in practical applications; Difficult to navigate. \\
\midrule
Free Text & Flexible; It allows natural expression of emotions. & Low agreement among annotators; Lower performance in practical applications; Lack of structure. \\
\bottomrule
\end{tabularx}
\end{table*}

In this work, the choice of selecting the appropriate scheme was challenging due to the plethora of work and debate regarding these schema. To find an appropriate scheme/classification of emotions to be used in this work, the following two main research objectives were set:

\begin{enumerate}
    \item \textbf{The emotions scheme should be at least statistically and scientifically valid, and not only empirical and observational.}
    
    The selected scheme needs to be derived from human annotators with sufficient statistical validation. This ensures that the scheme is based on reliable human judgment and is statistically sound. Additionally, it must be acceptable by the community and commonly used in various applications, ensuring its relevance and practical utility. Finally, the scheme should counter as many drawbacks of the existing schema as possible, addressing issues such as oversimplification and insufficient emotional coverage to provide a more comprehensive and nuanced understanding of emotions.

    \item \textbf{The emotions scheme must be appropriate and aligned with the concepts of LDA.} 
    
    This alignment is necessary to ensure that the emotional classifications contribute meaningfully to the discovery of Lacanian discourses within the text. LDA involves the study of language and its effects on the subject, emphasizing the importance of underlying structures and meanings in discourse. Therefore, the chosen scheme should be capable of capturing these nuances and complexities in emotional expression, facilitating a deeper understanding of the text in line with Lacanian principles.
\end{enumerate}

\subsection{Related Work}
\label{sec:relatedwork}
To the best of our knowledge, there isn't any published work that tried to identify a relationship between emotions and the Lacanian Discourses in a systematic way. There are theoretical conjectures \cite{soler2016lacanian} that propose some relationships  as for the \textbf{Master Discourse}:
    \begin{itemize}
        \item Anger - in reaction to authority or oppression.
        \item Fear - of punishment or reprisal.
        \item Resentment - toward those in power.
    \end{itemize}

Such conjectures remain to be validated both qualitatively and quantitatively as well. Qualitatively in the sense of verifying if the mentioned emotions are truly related to the associated discourse. Quantitatively in the sense to know how frequent the mentioned emotion appears in the associated discourse and what is its differentiating power.

In this section a few works are briefly reviewed not because they adopted similar approaches to this work and could be compared with but because of: (i) their background information that helps to understand the context in which this work has been developed; and (ii) their ideas that could be used to fuel future research.

In \cite{bucci2022}, it could appear at first sight that the authors deal exactly with what is done in this work. It is not so. The authors present two independent study cases. 

The first one is an Emotional Text Analysis (ETA). They show that emotions expressed in language are not individual phenomena, but organizers of historically determined social relations. They conducted focus groups with young graduates, who were about to start their first job in multinational corporations, and with their corporate mentors. Applying ETA they identified two independent and different clusters of emotions. One was associated to young graduates' expectations. The other was associated to the mentor's accounts. In this study case there isn't any mention to LDA.

The second study case is an example of LDA carried out within a therapeutic community (TC) for children and teenagers with a diagnosis of psychosis and autistic spectrum disorders. Conversations between the therapeutic group and the family group were transcribed verbatim and read by a panel of researchers trained in Lacanian psychoanalysis. The family group adopts a \textbf{Hysteric Discourse} and claims for the therapeutic group to take the \textbf{Master Discourse} and to provide answers and solution to the problem being dealt with. The leader of the therapeutic group, wisely does not fall into the trap and adopts the \textbf{Analyst Discourse} opening a path of further clarification of the problem.

This study case is closely related to this research and shows the possibility of applying LDA to daily dialogues and how the \textit{Agent} changes roles as the dialog evolves.

In another significant work \cite{vanheule2016}, initially the author makes a thorough revision of the four classic Lacanian Discourses. This revision is very important because it is not limited to the explanations given by J. Lacan himself at the time of his writings and other researchers that repeated what J. Lacan said in easier terms. For example, in the case of the \textbf{Master Discourse} besides reinforcing that it is a discourse associated with authority and power he includes:
\begin{quote}
    In the \textbf{Master Discourse} insistent signifiers ({\bf S1}) that provoke explanation ({\bf S2}) come to the fore. Their articulation connotes the subject underlying this articulation of signifiers ({\bf \$}), and at the same time provokes an excitement ({\bf a}) that is split off from the subject. 
\end{quote}

The author explains the importance of the changes:
\begin{itemize}
    \item \textit{Agent} $\rightarrow$ \textit{Semblance};
    \item \textit{other} $\rightarrow$ \textit{Jouissance};
    \item \textit{Product} $\rightarrow$ \textit{Surplus-jouissance};
\end{itemize}

Next, the author delves into the \textbf{Capitalist Discourse}. He clarifies that initially J. Lacan presented this discourse as just a variation of the \textbf{Master Discourse} and during sometime the true meaning of it was not properly understood. The \textbf{Capitalist Discourse} has never appeared in J. Lacan's seminars as a fifth discourse. Only in his lectures at the University of Milan \cite{contri2008} the structure of this discourse was formalized. The \textbf{Capitalist Discourse} is prevalent nowadays and cannot be left out of any work dealing with LDA and LDD. It is at the basis of the rupture of the social bonds and is the bedrock of many addictions that plague the society. It has clinical importance as it has to be managed differently if the patient is either a psychotic or a neurotic.

In an important work focusing on the role of signifiers, at the Introduction of \cite{bazan2023} the authors wrote:
\begin{quote}
    Freud proposed that names of clinically salient objects or situations, such as for example a beetle (\textit{Käfer}) in Mr. E’s panic attack, refer through their phonological word form, and not through their meaning, to etiologically important events—here, “\textit{Que faire?}” which summarizes the indecisiveness of Mr. E’s mother concerning her marriage with Mr. E’s father. Lacan formalized these ideas, attributing full-fledged mental effectiveness to the signifier, and summarized this as “\textit{the unconscious structured as a language}”. We tested one aspect of this theory, namely that there is an influence of the ambiguous phonological translation of the world upon our mental processing without us being aware of this influence.
\end{quote}

The aforementioned work is very recent, published in 2023, constitutes an empirical evidence for the mental effectiveness of the signifier. It validates the empirical approach adopted by our own work and highlights the importance and possibility of identifying signifiers and the corresponding signifying chain that is at the basis of LDA and LDD.

Another important line of research was developed by Robert “Rob” Haskell (1938-2010) who was Professor of psychology and department chair at the University of New England. He earned his PhD from the Pennsylvania State University in Psychology and Social Relations, and was a Charter member of the American Psychological Society. His areas of research include: transfer of learning, small group leadership, language and communication, \textbf{unconscious cognition} \cite{haskell2008deep}, and analogical reasoning. He developed a novel \textbf{logico-mathematic, structural methodology for the analysis and validation of sub-literal (SubLit) language and cognition} \cite{haskell2003logico}.

As a psychologist, not a psychoanalyst, Prof. Haskell always claimed that ``his'' unconscious is essentially different from Freud's \textit{Unconscious}. It out of the scope of this work to engage in such discussion. His work is quite important because it shows, by means of several examples and a solid theoretical basis, the presence of a hidden \textit{Truth} behind a discourse. To the best of our knowledge the algorithm proposed by him in \cite{haskell2003logico} has never been implemented and may be a valid dimension to be included in future research to automatically identify signifiers and the associate signifying chain.

\section{Adopted Methodology}
\label{sec:adopted_methodology}
The adopted methodology encompasses a number of decisions that are discussed and justified in this section. They are:

\begin{itemize}
    \item Choice of working only with texts.
    \item Choice of using dialogues.
    \item Choice of the emotions set.
    \item Emotions and discourses assignments, voting and dataset creation process.
    \item Number of dialogues.
    \item Number of voters.
    \item Common-user's vote criteria for emotions and discourses.
    \item Probabilistic formulae to evaluate the relation between discourses and emotions.
\end{itemize}

\subsection{Choice of the emotions set}
\label{sec:choice_of_emotion_set}
In this section, solid arguments for choosing a particular set of emotions that can potentially be combined and used with the discovery of the Five Lacanian Discourses, and why they are aligned with the fundamental aspects of the LDA and LDD are provided.

In the paper "Classifying Emotion: A Developmental Account" \cite{zinck_classifying_2006}, Alexandra Zinck and Albert Newen propose a systematic classification of emotions that accounts for their complexity and developmental stages. They distinguish between four developmental stages of emotions:

\begin{enumerate}
    \item \textbf{Pre-emotions}: These are unfocused expressive emotional states, primarily observed in infants and characterized as either generally positive or negative.
    \item \textbf{Basic emotions}: These emotions are innate and do not require cognitive processing. They include fear, anger, joy, and sadness.
    \item \textbf{Primary cognitive emotions}: These emotions involve minimal cognitive content and are extensions or modifications of basic emotions.
    \item \textbf{Secondary cognitive emotions}: These are highly complex emotions that\textit{ depend on cultural information and personal experience}. They develop within the four dimensions of the basic emotions and are enriched by cognitive mini-theories, resulting in more finely grained emotions.
\end{enumerate}

Secondary cognitive emotions are particularly important as they are dependent on cultural information and personal experience, making them relevant for a nuanced analysis of emotional expression in texts. \textit{Additionally, this category is correlated with and related to the Lacanian Symbolic order, which must be considered when approaching a text within the field of LDA} \cite{Frosh2013ch1}. This category also consists of the emotions used in the GoEmotions dataset, which was selected to be used in this work.

The GoEmotions dataset, developed in \cite{demszky_goemotions_2020}, is the largest manually annotated dataset consisting of 58,009 English Reddit comments labeled for 27 emotions and Neutral. This dataset was created by researchers at Google Research, including Alan Cowen, a pioneer in emotion research. It offers a fine-grained typology adaptable to multiple downstream tasks, such as building empathetic chatbots or detecting harmful online behavior. The high quality of the annotations is demonstrated via Principal Preserved Component Analysis (PPCA), showing reliable dissociation among the 27 emotion categories \cite{demszky_goemotions_2020}.

Aligned with the classification proposed by Zinck and Newen \cite{zinck_classifying_2006}, the emotions in the emotions in the GoEmotions dataset can be further broke down as follows:

\begin{itemize}
    \item \textbf{Pre-Emotions}: Comfort
    \item \textbf{Basic Emotions}: Sadness, Joy, Fear, Anger
    \item \textbf{Primary Cognitive Emotions}: Relief, Nervousness, Excitement, Disappointment, Annoyance, 
    \item \textbf{Secondary Cognitive Emotions}: Admiration, Approval, Caring, Confusion, Curiosity, Desire, Disapproval, Disgust, Embarrassment, Gratitude, Grief, Love, Optimism, Pride, Realization, Remorse, Surprise
\end{itemize}

The GoEmotions dataset is statistically valid, widely accepted, and suitable for various applications, including NLP. This extensive dataset is curated to provide a comprehensive and nuanced classification of emotions, aligning well with the developmental stages of emotions.

The robustness of the emotions used in the dataset is demonstrated through various statistical analyses. For example, the dataset's annotations were found to be highly reliable, with 94\% of examples having at least two raters agreeing on one label, and 31\% having three or more raters in agreement. The high quality of these annotations is further validated through Principal Preserved Component Analysis (PPCA), which shows a strong dissociation among the 27 emotions. This ensures that the emotions captured in the dataset are both distinct and representative of a wide range of human emotional experiences.

Furthermore, the GoEmotions dataset has proven its utility in various NLP applications. It provides a solid baseline for emotion classification models, particularly when fine-tuning models like BERT, achieving significant performance metrics. This makes it highly suitable for tasks such as building empathetic chatbots, analyzing customer feedback, and detecting harmful online behavior. The dataset's adaptability to multiple downstream tasks underscores its practical value and broad applicability.

In summary, the selection of this schema aligns perfectly with the two objectives stated in Section \ref{sec:classification_of_emotions}. Firstly, the GoEmotions dataset meets the requirement that the schema should be statistically and scientifically valid, and not only empirical and observational. The dataset’s high reliability, demonstrated through rigorous statistical validation, ensures that it is based on sound human judgment and is widely accepted in the community, fulfilling the first objective. Secondly, the selected schema is also appropriate and aligned with the concepts of LDA and LDD, as it captures the nuanced and complex emotional expressions necessary for meaningful discourse analysis in line with Lacanian principles. This ensures that this research work is both methodologically sound and theoretically aligned, facilitating a deeper understanding of emotions within the framework of LDA and LDD.

To complete the set of emotions for this work, two more emotions, ``anguish'' and ``anxiety'', were added to the dataset because they are highly referenced in the psychoanalytic literature (see \ref{sec:emotions_lacan} and \ref{sec:emotions_freud}).

\subsection{Choice of working only with texts}
\label{sec:work_only_text}
The current body of research lacks a comprehensive methodology for systematically identifying traces of Lacanian discourses across various modalities. While Parker's work \cite{Parker2010} on LDA in interview texts provides valuable insights into the identification of these discourses, it has not yet evolved into a fully developed framework for the systematic and efficient discovery of Lacanian discourses across different data types.

Due to the emerging nature of research in this field, this study focuses exclusively on textual data, rather than incorporating modalities such as voice, sound, or video. Although relying solely on text may limit the exploration and identification of certain nuances inherent in Lacanian discourses, it offers a solid foundation for developing a methodological approach. Textual analysis provides a clear and structured starting point that can be refined and expanded in future research to include additional modalities. By beginning with text, the intent is to establish a robust analytical framework that can serve as a basis for more comprehensive studies in the future, introducing additional multi-modalities as well.

\subsection{Choice of using dialogues}
\label{sec:choice_of_dialogoues}
As described in Section \ref{sec:lacanian_discourses}, Lacanian discourses consist of two integral components: the sender and the receiver of the discourse. To facilitate the exploration of potential emotions associated with these discourses and given the limited prior work in this area, it can be realized that textual dialogues are ideal for such examination. Dialogues reveal attributes and interpretative nuances that standalone or multi-paragraph texts do not provide.

Dialogues inherently simulate the foundational structure of Lacanian discourses, which involve communication between at least two parties conveying a message. As dialogues unfold, they reveal latent information and intentions of the interacting persons. This dynamic flow of information makes it easier to identify the elements that constitute the Lacanian discourses, i. e., the components  {\bf S1}, {\bf S2}, {\bf \$} and {\bf a}.

Considering the aforementioned factors, the open-source dataset DAILY DIALOG \cite{dailydialogs} has been selected, which comprises everyday dialogues between two speakers. While other open-source datasets, such as Friends, could be employed, as they have being written for specific purposes (e.g., screenplays intended for comedy or sarcasm), they might introduce bias and skew in the study's results. Although these datasets may be useful in future research, as has been demonstrated by \cite{Joshi2016} and \cite{Poria2019}, they do not align with the current research objectives.

The choice of this dataset was guided by the need for authentic, natural conversations that closely mirror real-life interactions. This authenticity is crucial for accurately analyzing the emotional and interpretative aspects of Lacanian discourses. Focusing on everyday dialogues is an attempt to ensure that the discovered patterns and attributes are representative of genuine communication rather than artificially constructed scenarios.

\subsection{Emotions and discourses assignments, voting and dataset creation process}
\label{sec:voting_creation_process}
To establish a theoretical correlation between Lacanian discourses and emotions, discourse and emotion annotations on dialogues were conducted using a custom-built platform. On this platform, each user was presented with a dialogue from the dataset and tasked with assigning discourses and emotions to each part of the dialogue. Along with these assignments, annotators provided the following metrics for each discourse or emotion:

\textbf{Confidence Score of Discourse:} This metric measures the level of certainty with which a particular discourse is assigned to a segment of the dialogue.

\begin{itemize}
    \item \textbf{Scoring System:} The confidence score ranges from \textit{Definitely Not}, \textit{Probably Not}, \textit{Probably Yes}, to \textit{Definitely Yes}, representing varying levels of assurance.
    \item \textbf{Purpose:} This score helps to gauge the reliability of the discourse assignment, ensuring that only strongly evidenced discourses receive higher confidence levels. By using this metric, it is possible to establish a clear and reliable mapping between dialogues and Lacanian discourses.
\end{itemize}

\textbf{Weight of Discourse:} This value, ranging from 0 to 1, represents the strength or potency of the discourse within the dialogue. A higher weight indicates a stronger presence and influence of the discourse.

\begin{itemize}
    \item \textbf{Significance:} A higher weight indicates a stronger presence and influence of the discourse. This metric provides insight into the significance of the discourse within the dialogue, assisting in prioritizing more influential discourses.
    \item \textbf{Application:} By evaluating the weight of each discourse, it is possible to identify the dominant discourses that shape the emotional and interpretative dynamics of the dialogue.
\end{itemize}

\textbf{Confidence Score of Emotion:} Similar to the confidence score of discourse, this metric measures the certainty of assigning a particular emotion to a segment of the dialogue.

\begin{itemize}
    \item \textbf{Scoring System:} It uses the same \textit{Definitely Not}, \textit{Probably Not}, \textit{Probably Yes}, to \textit{Definitely Yes} scale to indicate the level of confidence in the emotional assignment.
    \item \textbf{Purpose:} This score ensures that emotional annotations are supported by strong evidence, enhancing the accuracy and reliability of the emotional analysis.
\end{itemize}

The voting process on the platform is designed to rigorously annotate dialogues with Lacanian discourses and emotions, supported by confidence and weight metrics. This method enables a systematic analysis of how discourses and emotions are intertwined, providing a foundational basis for understanding their theoretical correlations. By utilizing a structured voting system, it is ensured that the annotations reflect both the presence and influence of discourses and emotions, contributing to a comprehensive exploration of Lacanian discourse theory in the context of emotional analysis. This annotation process lays the groundwork for statistical modeling and in-depth analysis to uncover potential correlations and patterns.

In Appendix \ref{sec:sc_platform}, a screenshot of the platform where the voters annotated the dialogues is presented.

\subsection{Probabilistic formulae to evaluate the relation between discourses and emotions}
\label{sec:formulas_eq}
To derive a valid mathematical formula for evaluating the results from the annotation process, regarding the relation among discourses and emotions, the following methodology was adopted.

Let \( S \) be the set of all sentences annotated. The \textbf{conditional probability} of occurrence of a set of \( n \) discourses \( d_1, d_2, \dots, d_n \) (where \( 1 \leq n \leq 5 \)) given a set of \( l \) emotions \( e_1, e_2, \dots, e_l \) (where \( 1 \leq l \leq {30} \)) in a sentence of the dataset is defined by Eq. (~\ref{eq:annotation_formula_main}):

\begin{figure*}[!h]
\centering
\begin{equation}
\text{Prob}(d_1, d_2, \dots, d_n \mid e_1, e_2, \dots, e_l) = \frac{\sum_{s \in S_{d_1, \dots, d_n, e_1, \dots, e_l}} \left(\prod_{i=1}^{n} c_{s}(d_i) \cdot \prod_{j=1}^{l} c_{s}(e_j)\right)}{\sum_{s \in S_{e_1, \dots, e_l}} \left(\prod_{i=1}^{k_s} c_{s}(d_i) \cdot \prod_{j=1}^{l} c_{s}(e_j)\right)}
\label{eq:annotation_formula_main}
\end{equation}
\end{figure*}
\FloatBarrier

\noindent{where:}
\begin{itemize}
    \item $c_{s}(d_i)$ is the \textbf{confidence score} assigned to the occurrence of the discourse $\bm{d}_{\bm{i}}$
    \item $c_{s}(e_j)$ is the \textbf{confidence score} assigned to the occurrence of the emotion $\bm{e}_{\bm{j}}$ in sentence \( s \) (where \( s \in S \)), for \( j = 1, 2, \dots, l \). Here, \( e_j \) (where \( 1 \leq j \leq 30 \)) represents one of the 30 emotions.
\end{itemize}

The numerator in the formula is a summation over a subset \( S_{d_1, \dots, d_n, e_1, \dots, e_l} \), which includes all sentences \( s\in S\) that contain the discourses \( d_1, \dots, d_n \) and the emotions \( e_1, \dots, e_l \). Within this sum, the product of the confidence scores \( c_{s}(d_i) \) for each discourse \( d_i \) and the confidence scores \( c_{s}(e_j) \) for each emotion \( e_j \) is included. This represents the aggregated confidence for instances where the discourses \( d_1, d_2, \dots, d_n \) and the emotions \( e_1, e_2, \dots, e_l \) co-occur within the dataset.

The denominator sums over a subset \( S_{e_1, \dots, e_l} \), which includes all sentences \( s\in S\) that contain the emotions \( e_1, \dots, e_l \), without any requirement on co-occurring discourses in the sentences. The denominator involves a product of confidence scores, but here the set of discourses \( d_i \) may vary, as indicated by the product \( \prod_{i=1}^{k_s} \); where \(k_s\) is the number of discourses in the sentence  \( s \). This part of the formula represents the total aggregated confidence for all occurrences of the emotions \( e_1, e_2, \dots, e_l \), regardless of the number of discourses they appear with, i.e, not only when the examined discourses \( d_1 \dots d_n \) occur. The denominator, therefore, captures the overall likelihood of the emotions occurring within the dataset, independently of the specific discourses associated with them.

This general formula can be adapted to calculate the probability for any number of discourses n (\( 1 \leq n \leq 5 \)) given any number of emotions l (\( 1 \leq l \leq 30 \)). By extending the summation and product operations accordingly, the method remains applicable whether one is evaluating the appearance of a single discourse, a set of multiple discourses, together with a combination of several emotions. The framework is flexible, allowing for the inclusion of more complex discourse-emotion relationships within the dataset, ensuring comprehensive analysis across different scenarios.

For example, the general formula of Eq. (\ref{eq:annotation_formula_main}) can be applied to any specific case involving different combinations of discourses and emotions, as shown below:

\begin{itemize}
    \item \textbf{Example 1: One Discourse, One Emotion:} 
    \[
    \text{Prob}(d_1|e_1) = \frac{\sum_{s \in S_{d_1,e_1}} {c_s}(d_1) \cdot {c_s}(e_1)}{\sum_{s \in S_{e_1}} \left(\prod_{i=1}^{k_s} {c_s}(d_i)\right) \cdot {c_s}(e_1)}
    \]

    Here, the numerator sums the product of the confidence scores for discourse \( d_1 \) given emotion \( e_1 \) across all sentences. The denominator sums the product of the confidence scores for all occurrences of emotion \( e_1 \) with various discourses.

    \item \textbf{Example 2: Two Discourses, One Emotion:} 
    \[
    \text{Prob}(d_1, d_2 \mid e_1) = \frac{\sum_{s \in S_{d_1 d_2,e_1}} {c_s}(d_1) \cdot c(d_2) \cdot {c_s}(e_1)}{\sum_{s \in S_{e_1}} \left(\prod_{i=1}^{k_s} {c_s}(d_i)\right) \cdot {c_s}(e_1)}
    \]
    Here, the numerator sums the product of the confidence scores for discourses \( d_1 \) and \( d_2 \) given emotion \( e_1 \) across all sentences. The denominator sums the product of the confidence scores for all occurrences of emotion \( e_1 \) with the discourses \( d_1 \) and \( d_2 \).

    \item \textbf{Example 3: Two Discourses, Two Emotions:} 
    \[
    \text{Prob}(d_1, d_2|e_1, e_2) = \frac{\sum_{s \in S_{d_1 d_2,e_1 e_2}} {c_s}(d_1) \cdot {c_s}(d_2) \cdot {c_s}(e_1) \cdot {c_s}(e_2)}{\sum_{s \in S_{e_1 e_2}} \left(\prod_{i=1}^{k_s} {c_s}(d_i)\right) \cdot {c_s}(e_1) \cdot {c_s}(e_2)}
    \]

    In this case, the numerator sums the product of the confidence scores for discourses \( d_1 \) and \( d_2 \) given emotions \( e_1 \) and \( e_2 \) across all sentences. The denominator sums the product of the confidence scores for all occurrences of the emotions \( e_1 \) and \( e_2 \) with the discourses \( d_1 \) and \( d_2 \).
\end{itemize}

A justification for the above complex probability formulation is needed. Indeed, a simpler formulation could have been adopted, where the probability would be the fraction of the co-occurrence of the examined discourse-emotion combinations over the occurrences of the considered emotions. Such an approach however could overlook crucial factors, including annotation confidence for discourses, discourse strength, and the confidence level for the associated emotions. An additional important factor that needs to be addressed is the ``discourse weight'' \( w_{di} \), which captures the ``strength'' of the occurrence of the discourse in a sentence, i.e., how prominently the discourse occurs in the sentence. The inclusion of these three factors provides additional insight into the co-occurrence of the discourse with the emotions. 

Table \ref{tab:low_but_high} is an example that shows the confidence scores for discourse \( d_1 \), the weight of discourse \( d_1 \), and the confidence scores for \( e_1 \) in a hypothetical dataset, where \( d_1 \) and \( e_1 \) appear together in only four sentences.

\begin{table*}[t]
\centering
\caption{An example showing the confidence scores for discourse \( d_1 \), the weight of discourse \( d_1 \), and the confidence scores for \( e_1 \) in a hypothetical dataset, where \( d_1 \) and \( e_1 \) appear together in only four sentences.}
\label{tab:low_but_high}

\begin{tabular}{|l|c|c|c|}
\hline
\textbf{Sentence} & \textbf{conf (d1)} & \textbf{conf (e1)} & \textbf{weight (d1)} \\
\hline
Sentence 1 & 0.2 & 0.4 & 1 \\
\hline
Sentence 2 & 0.3 & 0.2 & 0.8 \\
\hline
Sentence 3 & 0.3 & 0.3 & 0.7 \\
\hline
Sentence 4 & 0.2 & 0.2 & 0.4 \\
\hline
\end{tabular}

\end{table*}

For the data in Table \ref{tab:low_but_high}, even when confidence levels are low, the probability remains high due to the limited annotated diversity in the small dataset, leading to an almost exclusive correlation. Specifically, as \( e_1 \) only appears with \( d_1 \) the conditional probability \( \text{Prob}(d_1|e_1) \) is as follows:

\[
\text{Prob}(d_1|e_1) = \frac{(0.2 \times 0.4) + (0.3 \times 0.2) + (0.3 \times 0.3) + (0.2 \times 0.2)}{(0.2 \times 0.4) + (0.3 \times 0.2) + (0.3 \times 0.3) + (0.2 \times 0.2)} = 1
\]

In other words, even this sophisticated probability definition falls short to model well the relationship between emotions and discourses. Therefore, it is indeed necessary to consider the strength of the discourse's presence. This is achieved by incorporating the weights of the discourses as described by Definition \ref{def:weight_level_def}

\begin{definition}[Weight Level]
\label{def:weight_level_def}
The \textbf{weight level (W)} of the co-occurrence of the discourses \( d_1, d_2, \dots, d_n \) with emotions \( e_1, e_2, \dots, e_l \) is defined as follows:
\begin{itemize}
    \item[i)] First, for each sentence \( s \) where this combination of discourses and emotions occurs, the product of the involved discourse weights \( w_s(d_i) \) (where \( 1 \leq i \leq n \)) is taken.
    \item[ii)] Then, to evaluate the total discourse weight level, the above product is summed over all relevant sentences \( s \) where the examined combination occurs.
\end{itemize}
The weight level is then given by:
\[
\text{W}(d_1, d_2, \dots, d_n, e_1, e_2, \dots, e_l) = \sum_{s \in S_{d_1 \dots d_n, e_1, \dots, e_l}} \prod_{i=1}^{n} w(d_i)
\]
\begin{flushright}
${_\square}$
\end{flushright}
\end{definition}

The relation R among discourses \( d_1, d_2, \dots d_n \) and emotions \( e_1,e_2, \dots, e_l \) is given in Definition \ref{def:relation}

\begin{definition}[Relation of Co-occurrence]
\label{def:relation}
The \textbf{relation} of the co-occurrence of the discourses \( d_1, d_2, \dots, d_n \) with emotions \( e_1, e_2, \dots, e_l \) is defined by Eq. (\ref{eq:final_relation}), as follows:
\begin{equation}
\label{eq:final_relation}
\text{R}(d_1, d_2, \dots, d_n, e_1, e_2, \dots, e_l) = \text{Prob}(d_1, d_2, \dots, d_n \mid e_1, e_2, \dots, e_l) \times \text{W}(d_1, d_2, \dots, d_n, e_1, e_2, \dots, e_l)
\end{equation}
\begin{flushright}
${_\square}$
\end{flushright}
\end{definition}


Using the data provided in Table \ref{tab:low_but_high}, the weight level and the relation between \( d_1 \) and \( e_1 \) are calculated as

\[
W(d_1 | e_1) = 1 \times 0.8 \times 0.7 \times0.4 = 0.224
\]

\noindent{and}

\[
R(d_1 | e_1) = Pr(d_1 | e_1) \times W(d_1 | e_1) = 1 \times 0.224 = 0.224
\]

Interestingly, the discourse-emotion relationship now admits fine-grained values, possibly covering a very broad spectrum regardless of the dataset size, since  all factors of the discourse-emotion relationship are now directly taken into account.

This result is then normalized, ensuring that the final value lies within a standard range of 0 to 1 where the maximum value is taken from all relations where the number of discourses is \( n \) and the number of emotions \( l \). This final relation is named the \textbf{relation intensity (RI)} and described by Definition \ref{def:normalized_relation}, for the co-occurrence of a certain combination of discourses \( d_1, d_2, \dots, d_n \) with certain emotions \( e_1, e_2, \dots, e_l \) and given by Eq. (\ref{eq:final_relation}):

\begin{definition}[Normalized Relation Intensity]
\label{def:normalized_relation}
The \textbf{normalized relation intensity} among discourses \( d_1, d_2, \dots, d_n \) and emotions \( e_1, e_2, \dots, e_l \) is defined as follows:
\begin{equation}
\label{eq:final_normalized_relation}
\text{RI}(d_1, d_2, \dots, d_n, e_1, e_2, \dots, e_l) = \frac{\text{R}(d_1, d_2, \dots, d_n, e_1, e_2, \dots, e_l)}{\max_{n,l} \left(\text{R}(d_{i1}, d_{i2}, \dots, d_{in}, e_{j1}, e_{j2}, \dots, e_{jl})\right)}
\end{equation}
\begin{flushright}
    ${_\square}$
\end{flushright}
\end{definition}

\section{Results and Discussion}
\label{sec:results}
In this section the experimental procedure, the processing of assignments (from now on designated as votes), the results along the discussion of them are detailed.

\subsection{Experimental procedure}

\begin{itemize}
    \item 40 dialogues were randomly selected from the database corresponding to 276 sentences.

    \item Three voters, familiar with the Lacanian Theory of Discourses, independently from each other, for each sentence:
    \begin{itemize}
        \item elected emotions from the pre-defined set, and assigned a confidence score;
        \item elected the appropriate discourses and assigned both a confidence score and a weight as described in Section \ref{sec:voting_creation_process}
    \end{itemize}

    \item The assignments of the three voters were processed and combined as the votes of a so called ``common user'' for both the assigned discourses and emotions as well.  
\end{itemize}

The pseudocode for the algorithm to derive the final discourses for the sentences is shown in Algorithm~\ref{alg:final_discourses} and its simplified version is shown in Table \ref{tab:disc_algo}.

The algorithm to derive to the discourses for the ``common user's'' vote considers only up to 4 discourses as this was the maximum number a voter assigned to a sentence. Also, the cases which led to discourse as ``none'' were discarded from the final analysis of the data.

\vspace{2mm}

Hereafter an example of dialog processing is shown to help the understanding the ``common user's'' vote evaluation.

\textit{Dialog ID:} $FroLatteAd\_10$

{\bf I can't believe Mr. Fro didn't buy it. Who does that guy think he is anyway? Bill Gates?}
\begin{itemize}
    \item Voter 1: Hysteric, High, 0.9;
    \item Voter 2: Hysteric, Mid, 0.7;
    \item Voter 3: Hysteric, High, 1.0;
    \item Common user -- Rule 1: Hysteric, High, 1.0.
\end{itemize}

{\bf He had a lot of nerve telling us our ads sucked.}
\begin{itemize}
    \item Voter 1: University, Mid, 0.6; Hysteric, Low, 0.2;
    \item Voter 2: Master, Low, 0.1; Hysteric, Mid, 0.6;
    \item Voter 3: Hysteric, High, 1.0;
    \item Common user -- Rule 6: Hysteric, High, 0.6.
\end{itemize}

{\bf Time to order. Balista, today I want a skinny triple latte.}
\begin{itemize}
    \item Voter 1: Hysteric, Mid, 0.4;
    \item Voter 2: Hysteric, Low, None; Capitalist, Low, 0.5;
    \item Voter 3: Hysteric, High, 0.8;
    \item Common user -- Rule 4: Hysteric, High, 0.8.
\end{itemize}

{\bf When did you start worrying about your weight?}
\begin{itemize}
    \item Voter 1: Analyst, High, 0.8;
    \item Voter 2: Analyst, Mid, 0.7;
    \item Voter 3: Analyst, High, 0.5; Hysteric, High, 0.5.
    \item Common user -- Rule 4: Analyst, High, 0.8.
\end{itemize}

{\bf I'm not. I just don't feel like drinking whole milk today. Why? Do you think I'm fat?}
\begin{itemize}
    \item Voter 1: Master, Mid, 0.5; Hysteric, High, 0.8;
    \item Voter 2: Master, Low, 0.2; Hysteric, Mid, 0.7;
    \item Voter 3: Master, High, 0.5; Hysteric, High, 0.5;
    \item Common user -- Rule 2: Master, High, 1.0; Hysteric, High, 1.0.
\end{itemize}

{\bf No, Jess, chill out!}
\begin{itemize}
    \item Voter 1: Master, Low, 0.2;
    \item Voter 2: Master, Low, 0.6; Hysteric, Low, 0.1;
    \item Voter 3: Master, High, 0.8;
    \item Common user -- Rule 4: Master, High, 0.8.
\end{itemize}

The example shows that:
\begin{itemize}
    \item The algorithm harmonizes the discrepancies among the individual voters.
    \item The harmonization is accomplished by discarding votes that are given just by one voter, and decreasing the value of the confidence score and the weight level in cases where the votes are not unanimous.
    \item The confidence score and weight level assigned by the individual voters are not taken into account in the derivation of the ``common user's'' vote. Nevertheless, these individual assignments are important when deriving the model for the individual voters. However, such modelling is out of the scope of the current work.
    \item The algorithm enables to capture the presence of both unique discourses and traces of multiple discourses in each of the statements.
\end{itemize}

The algorithm to derive to the emotions for the ``common user's'' vote is shown in the Algorithm \ref{alg:emotion_confidence}

\begin{table*}[t]
\centering
\caption{Summary of rules to derive the discourses assignment of the ``common user''}
\label{tab:disc_algo}
\begin{tabular}{|l|l|l|l|l|}
\hline
\textbf{Rules} & \textbf{Voter 1} & \textbf{Voter 2} & \textbf{Voter 3} & \textbf{Outcome (discourse, confidence score, weight)} \\
\hline
Rule 1 & $d_1$ & $d_1$ & $d_1$ & $(d_1,\text{H},1)$ \\
\hline
Rule 2 & $d_1\ d_2$ & $d_1\ d_2$ & $d_1\ d_2$ & $(d_1,\text{H},1)\ (d_2,\text{H},1)$ \\
\hline
Rule 3 & $d_1\ d_2$ & $d_1\ d_2$ & $d_1$ & $(d_1,\text{H},1)\ (d_2,\text{M},1)$ \\
\hline
Rule 4 & $d_1\ d_2$ & $d_1$ & $d_1$ & $(d_1,\text{H},0.8)$ \\
\hline
Rule 5 & $d_1\ d_2\ d_3$ & $d_1$ & $d_1$ & $(d_1,\text{H},0.6)$ \\
\hline
Rule 6 & $d_1\ d_2$ & $d_1\ d_3$ & $d_1$ & $(d_1,\text{H},0.6)$ \\
\hline
Rule 7 & $d_1\ d_2\ d_3$ & $d_1\ d_2$ & $d_1\ d_2$ & $(d_1,\text{H},0.8)\ (d_2,\text{H},0.8)$ \\
\hline
Rule 8 & $d_1$ & $d_1$ & (none) & $(d_1,\text{M},1)$ \\
\hline
Rule 9 & $d_1$ & $d_1$ & $d_2$ & $(d_1,\text{M},1)$ \\
\hline
Rule 10 & $d_1$ & $d_1\ d_3$ & $d_2$ & $(d_1,\text{M},0.8)$ \\
\hline
Rule 11 & $d_1\ d_2$ & $d_2\ d_3$ & $d_1\ d_2$ & $(d_1,\text{M},1)\ (d_2,\text{H},0.8)$ \\
\hline
Rule 12 & $d_1\ d_2$ & $d_1$ & $d_2$ & $(d_1,\text{M},1)\ (d_2,\text{M},1)$ \\
\hline
Rule 13 & $d_1\ d_2$ & $d_1\ d_3$ & (none) & $(d_1,\text{M},0.6)$ \\
\hline
Rule 14 & $d_1\ d_2$ & $d_1\ d_2$ & (none) & $(d_1,\text{M},1)\ (d_2,\text{M},1)$ \\
\hline
Rule 15 & $d_1\ d_2$ & $d_1\ d_2$ & $d_3\ d_4$ & $(d_1,\text{M},1)\ (d_2,\text{M},1)$ \\
\hline
Rule 16 & $d_1$ & $d_1$ & $d_3\ d_4$ & $(d_1,\text{M},1)$ \\
\hline
Rule 17 & $d_1$ & $d_1\ d_2$ & $d_2$ & $(d_1,\text{M},1)\ (d_2,\text{M},1)$ \\
\hline
Rule 18 & $d_1\ d_4$ & $d_2\ d_3\ d_4$ & $d_1\ d_2\ d_3$ & $(d_1,\text{M},1)\ (d_2,\text{M},1)\ (d_3,\text{M},1)\ (d_4,\text{M},1)$ \\
\hline
Rule 19 & $d_1$ & $d_1\ d_2$ & $d_2\ d_3$ & $(d_1,\text{M},1)\ (d_2,\text{M},0.8)$ \\
\hline
Rule 20 & $d_1$ & $d_2\ d_3\ d_4$ & $d_2$ & $(d_2,\text{M},0.6)$ \\
\hline
Rule 21 & (none) & $d_1\ d_2$ & (none) & $(\text{none},\text{L},0)$ \\
\hline
Rule 22 & $d_1\ d_2$ & $d_1\ d_2$ & $d_1\ d_3$ & $(d_1,\text{H},0.8)\ (d_2,\text{M},1)$ \\
\hline
Rule 23 & $d_1$ & (none) & (none) & $(\text{none},\text{L},0)$ \\
\hline
Rule 24 & $d_1\ d_2$ & $d_1$ & $d_3$ & $(d_1,\text{M},0.8)$ \\
\hline
Rule 25 & $d_1\ d_2$ & $d_1\ d_2$ & $d_3$ & $(d_1,\text{M},1)\ (d_2,\text{M},1)$ \\
\hline
Rule 26 & $d_1$ & \makecell[l]{$d_1$\\$d_2$} & \makecell[l]{$d_1$\\$d_2$\\$d_3$} & $(d_1,\text{H},0.8)\ (d_2,\text{M},0.8)$ \\
\hline
Rule 27 & $d_1\ d_2$ & $d_1$ & (none) & $(d_1,\text{M},0.8)$ \\
\hline
Rule 28 & $d_1\ d_2$ & $d_1\ d_3$ & $d_2\ d_3$ & $(d_1,\text{M},1)\ (d_2,\text{M},1)\ (d_3,\text{M},1)$ \\
\hline
Rule 29 & \makecell[l]{$d_1$\\$d_2$} & \makecell[l]{$d_1$\\$d_3$} & $d_2$ & $(d_1,\text{M},0.8)\ (d_2,\text{M},1)$ \\
\hline
\end{tabular}
\end{table*}

\subsection{Votes processing}
\label{sec:votes processing}

Eq. (\ref{eq:annotation_formula_main}) is applied to the ``common user's'' votes and a complete table with all conditional probabilities is generated. Table \ref{tab:cond_probs} shows an extract of the complete table of conditional probabilities. In any given row, the first column is a set of up to three simultaneous emotions and the remaining columns show the conditional probability of the set of the discourses. It may be seen that the sum of the rows probabilities add to 1. In the complete table the sum of the columns probabilities also add to 1.

\vspace{3mm}

\begin{table*}[t]
\centering
\caption{Conditional Probabilities for Emotional Sets (M = Master, U = University, A = Analyst, H = Hysteric, C = Capitalist)}
\label{tab:cond_probs}

\resizebox{\textwidth}{!}{ 
\begin{tabular}{|l|c|c|c|c|c|c|c|c|c|c|c|c|c|c|}
\hline
\textbf{Emotion} & \textbf{A, C} & \textbf{A, H} & \textbf{A, H, M} & \textbf{A, M} & \textbf{C, H} & \textbf{H, M} & \multicolumn{2}{c|}{\textbf{H, M, U}} & \textbf{M, U} & \textbf{A} & \textbf{C} & \textbf{H} & \textbf{M} & \textbf{U} \\
\hline
('annoyance', 'disappointment', 'disapproval') & 0 & 0.0884 & 0 & 0 & 0 & 0.6389 & \multicolumn{2}{c|}{0.075953} & 0 & 0 & 0 & 0.1136 & 0.0830 & 0 \\
\hline
('admiration', 'approval', 'excitement') & 0 & 0 & 0 & 0 & 0 & 0 & \multicolumn{2}{c|}{0.049467} & 0 & 0 & 0 & 0.8617 & 0 & 0 \\
\hline
('anger', 'annoyance', 'disapproval') & 0 & 0 & 0 & 0 & 0 & 0.0491 & \multicolumn{2}{c|}{0} & 0 & 0 & 0 & 0.7052 & 0.1146 & 0.1310 \\
\hline
('admiration', 'approval', 'desire') & 0.1926 & 0 & 0 & 0 & 0 & 0 & \multicolumn{2}{c|}{0} & 0 & 0 & 0 & 0.8074 & 0 & 0 \\
\hline
('annoyance', 'disapproval', 'realization') & 0 & 0 & 0 & 0 & 0 & 0 & \multicolumn{2}{c|}{0.5453} & 0 & 0 & 0 & 0.4547 & 0 & 0 \\
\hline
\end{tabular}
}
\end{table*}

\vspace{3mm}

Next, Eq. (\ref{eq:final_normalized_relation}) is applied to the conditional probabilities generating a complete table of the relation intensity between a set of emotions and a set of discourses. In any given row, the first column is a set of up to three simultaneous emotions and the remaining columns show the value of the relation intensity for all 13 discourses combinations that were considered. 

For clarity, we symbolize the relation intensity for particular combinations of discourses with emotions 
\[
\mathbf{RI}(d_1, d_2, \dots, d_n, e_1, e_2, \dots, e_l)
\]
in the following way: 
\[
\{ e_1, e_2, \dots, e_l \} \, \mathbf{RI} \, \{ d_1, d_2, \dots, d_n \},
\]
\noindent{making it easier the identification of the involved emotions and discourses.}

In general terms, this relation intensity can be represented as:

\[
\{e_i, e_j, e_k\} \mathbf{RI} \{d_a, d_b, d_c\} = r_m,
\]

\noindent{where $0 \leq \{i, j, k\} \leq 30$}. The value 0 is used to represent ``no emotion considered'' and the other values point to a specific emotion of the dataset.

$\{a, b, c\}$ represent one of the five Lacanian discourses or none of them to take into account single or combination of discourses.

$1 \leq m \leq 13$ represents one of the 13 possible discourses combinations and $0 \leq r_m \leq 1$. As already shown by Eq. (\ref{eq:final_normalized_relation}),  $r_m$ is obtained by the multiplication of a probability and a weight, both numbers are less or equal to 1, and a further normalization. So, it is not possible to have a high value of the relation intensity if the corresponding conditional probability is low. For each set of emotions $\{e_i, e_j, e_k\}$ the 13 values of $r_m$ can be sorted in descending order indicating the most prevalent discourses associated to that set of emotions.

\begin{figure}[p]
    \centering
    \includegraphics[width=0.45\textwidth]{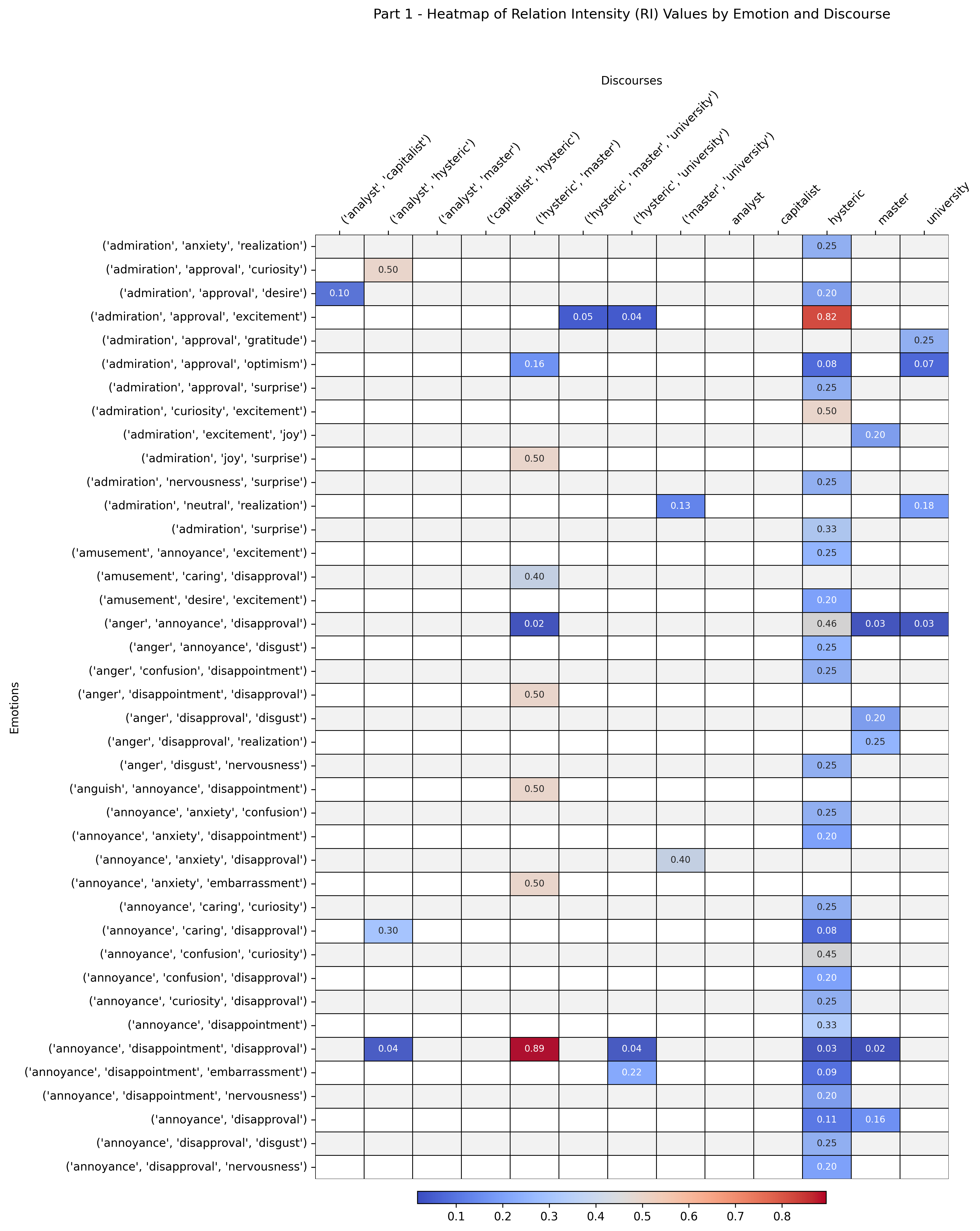}
    \hspace{0.05\textwidth} 
    \includegraphics[width=0.45\textwidth]{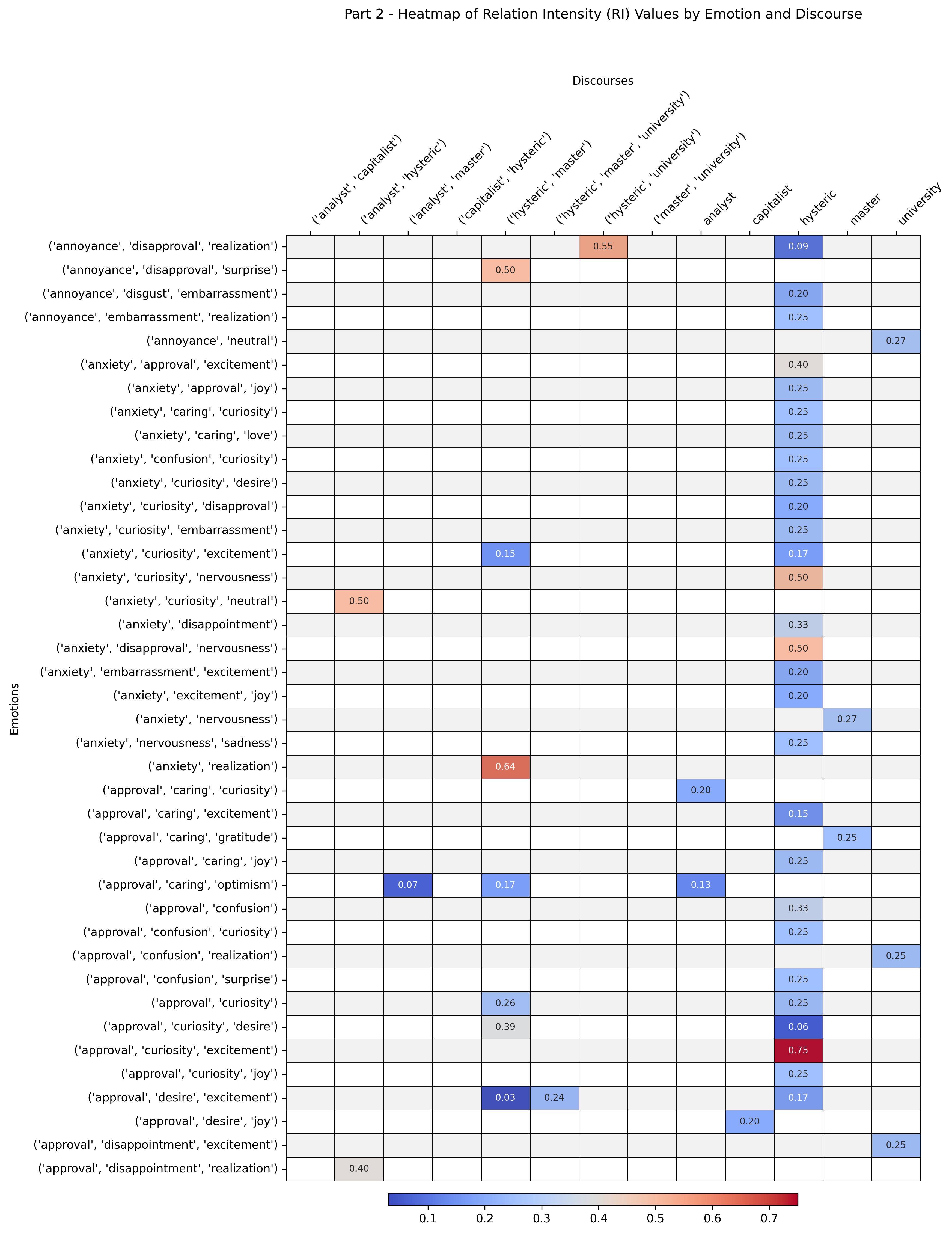}

    \vspace{\baselineskip} 

    \includegraphics[width=0.45\textwidth]{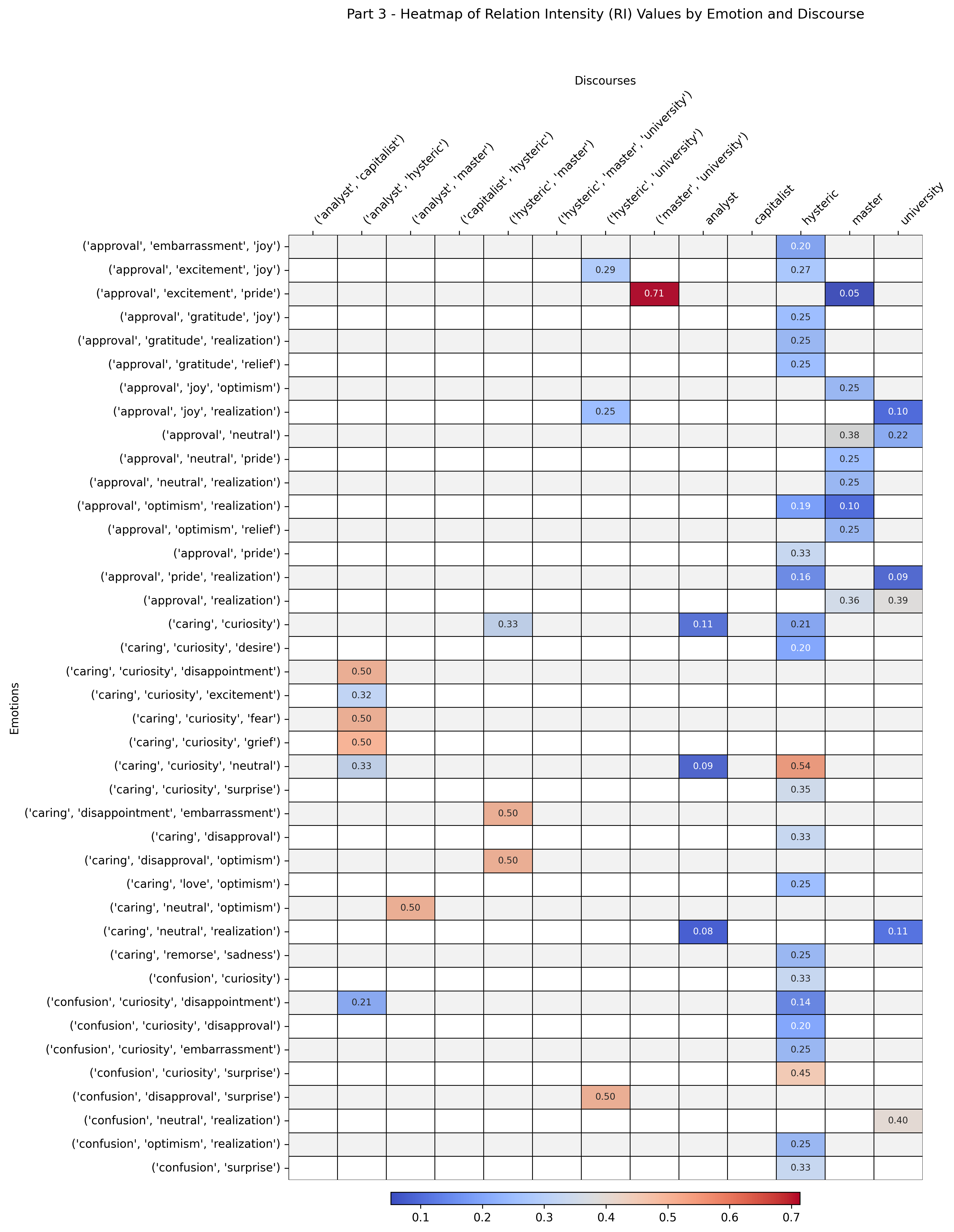}
    \hspace{0.05\textwidth} 
    \includegraphics[width=0.45\textwidth]{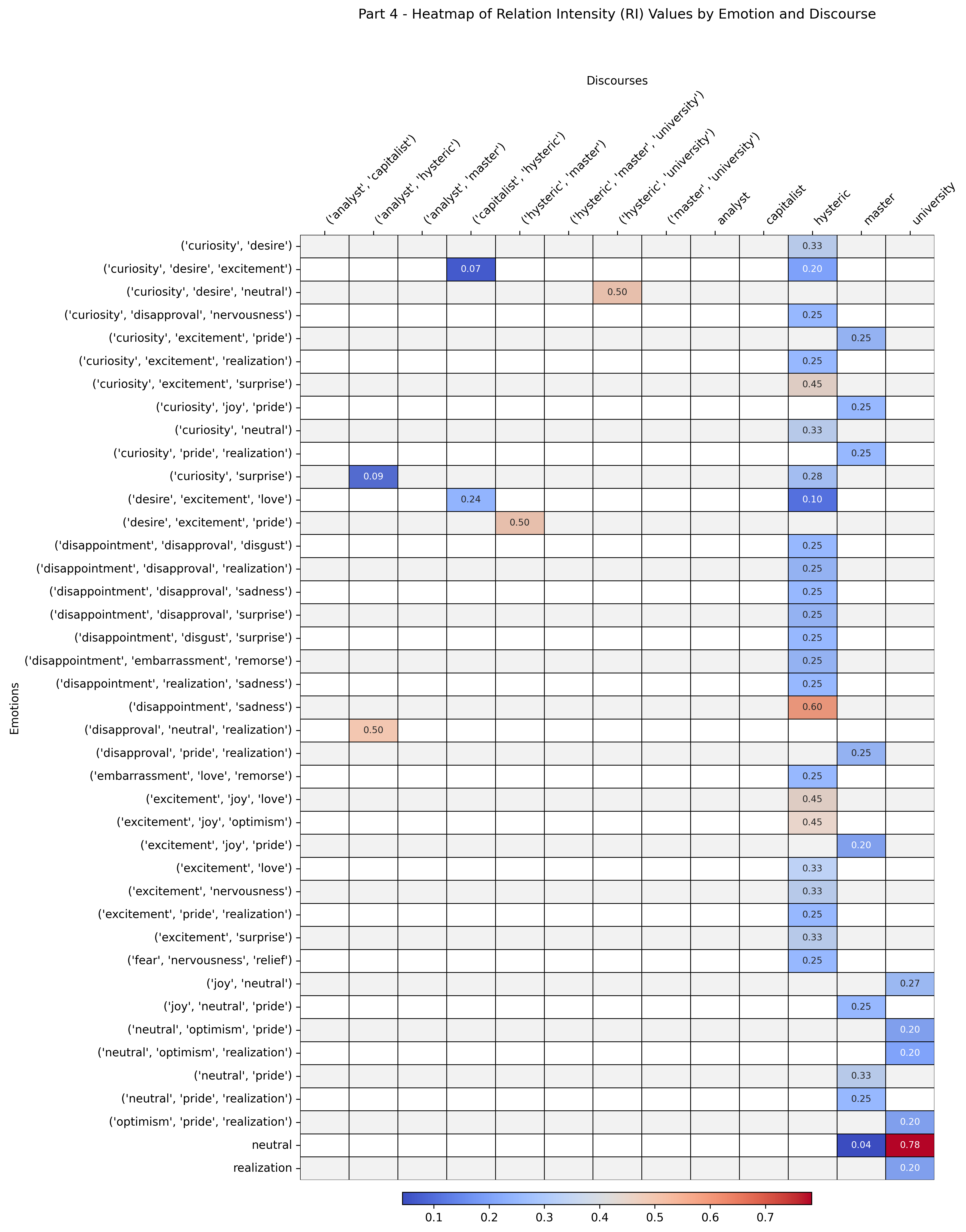}

    \caption{The complete heat map of the relation intensity between emotions and Lacanian discourses displayed in a 2x2 grid.}
    \label{fig:four_figures}
\end{figure}

The relation intensity results have been in summarized in a heat map shown in Figure \ref{fig:four_figures}, that is limited to the top 5 values of the relation intensity for each combination of emotions. The complete relation intensity table has 198 rows while top 5 values limited table has only 67 rows.
These heat maps show the entire dataset and are not limited to just the top 5.

\vspace{3mm}

\subsection{Findings and discussion}
It is important to point out that the findings and corresponding explanations are preliminary due to the limitations of the experimental procedure and decisions taken for the adopted methodology. Nevertheless we are confident that the qualitative findings are of significance and the quantitative ones deserve to be refined in future work.

The heat map shown in \ref{fig:four_figures} must be read row-wise meaning that comparisons of values in the same row are meaningful. However, comparison of values in different rows have a qualitative meaning but not a quantitative one. In the sequence, some examples will clarify these comments.

Out of the 67 rows of the heat map \textbf{only 7 have more than one value different than zero}. This result means that the identification of emotions has a very strong differential power to identify the prevalent discourse in a sentence. The rows that have more than one relation intensity value different than zero are the following:

\begin{itemize}   
    \item \{admiration, approval, excitement\} $\mathbf{RI}$ \{M, H, U\} = 0.05;
    \item \{admiration, approval, excitement\} $\mathbf{RI}$ \{H\} = 0.82.

    It can be seem that the Hysteric discourse is identified in both instances. The relation intensity for the only Hysteric discourse is very strong. This is not surprising as the excitement is in both set of emotions. The presence of the Hysteric discourse along the Master and University may also happen and this identification is certainly due to the \textit{judgement} emotions -- admiration and approval.
\end{itemize}

\vspace{3mm}

\begin{itemize}   
    \item \{approval, caring, optimism\} $\mathbf{RI}$ \{M, A\} = 0.07;
    \item \{approval, caring, optimism\} $\mathbf{RI}$ \{A\} = 0.13.
    
    It is reasonable to find the feeling of caring associated to the Analyst discourse. In a pure analyst position neither any kind of \textit{judgement} nor \textit{expectation} should appear, so the presence of approval and optimism counts for the identification of the Master discourse as well.
\end{itemize}

\vspace{3mm}

\begin{itemize}   
    \item \{approval, neutral\} $\mathbf{RI}$ \{M\} = 0.38;
    \item \{approval, neutral\} $\mathbf{RI}$ \{U\} = 0.22.
    
\end{itemize}

and,

\begin{itemize}   
    \item \{approval, realization\} $\mathbf{RI}$ \{M\} = 0.36;
    \item \{approval, realization\} $\mathbf{RI}$ \{U\} = 0.39.
    
   In both cases there is an ambiguity between the Master and the University discourses. The approval emotion is always present, i.e., the characteristic of \textit{judgement} associated to these discourses. The other two emotions, neutral and realization, indicate the characteristic of objectivity of these discourses. It is reasonable that such ambiguity occurs once {\bf S1} is, in both cases, on the transmitter side, either as the \textit{Agent} or the hidden \textit{Truth}. In addition, {\bf S1} either drives {\bf S2} or addresses it.
\end{itemize}

\vspace{3mm}

\begin{itemize}   
    \item \{caring, curiosity\} $\mathbf{RI}$ \{M, H\} = 0.33;
    \item \{caring, curiosity\} $\mathbf{RI}$ \{A\} = 0.11.
\end{itemize}

and,

\begin{itemize}   
    \item \{caring, curiosity, neutral\} $\mathbf{RI}$ \{H\} = 0.54;
    \item \{caring, curiosity, neutral\} $\mathbf{RI}$ \{A\} = 0.09.
    
    In all cases the emotions caring and curiosity are present. They are distinctively associated to the Analyst discourse but in a subtle and elusive way. This is probably the reason why the relation intensity has low values. However, caring and curiosity are not exclusive to the Analyst discourse. In the case of the Hysteric discourse the speaker may look intensively, emotionally charged, for objective information (neutrality). In the case of the Master discourse the speaker may also show curiosity and care about the potential production of the receiver.
\end{itemize}

 \vspace{3mm}
 
\begin{itemize}   
    \item \{neutral\} $\mathbf{RI}$ \{M\} = 0.04;
    \item \{neutral\} $\mathbf{RI}$ \{U\} = 0.78.
    
   Neutrality is expected from the speaker providing objective information as is the case of the University discourse. In a much lesser degree neutrality may characterize a Master discourse when it describes a scenario in which the receiver is expecting to be acting. Neutrality may be less intense in this case because the Master discourse is also characterized by authority, power and self-image which may imply a biased speech.
\end{itemize}

The \textbf{only Master discourse} is uniquely identified by a very large set of emotions combinations, 10 in total. The relation intensity is in the range:

\[
0.25 \leq \{e_i, e_j, e_k\} \mathbf{RI} \{M\} \leq 0.38,
\]

\noindent{and the emotions show a prevalence of judgement (approval, disapproval), self-image (pride),  jouissance (joy) and expectation (anxiety, nervousness). This finding is consistent with {\bf S1} in the position of \textit{Agent / Semblance} in which in a role of power and authority may produce statements charged with judgement and expectation. On the other hand, {\bf S1} driven by {\bf \$} speaks about himself in terms of the image he/she wants to display.}

\textbf{The top 4 emotions combinations in relation to the Master discourse are:}
\begin{itemize}
    \item \{approval, neutral\} $\mathbf{RI}$ \{M\} = 0.38;
     \item \{approval, realization\} $\mathbf{RI}$ \{M\} = 0.36;
    \item \{neutral, pride\} $\mathbf{RI}$ \{M\} = 0.33;
    \item \{anxiety, nervousness\} $\mathbf{RI}$ \{M\} = 0.27.  
\end{itemize}

Several other emotions combinations present the same relation intensity equal to 0.25.

The \textbf{only Analyst discourse} is uniquely identified by a set of 5 emotions combinations. The relation intensity is in the range:

\[
0.08 \leq \{e_i, e_j, e_k\} \mathbf{RI} \{A\} \leq 0.20,
\]

\noindent{and the emotions show a prevalence of caring associated with curiosity and neutrality. This finding is consistent with {\bf a} in the position of \textit{Agent / Semblance} driven by {\bf S2} in a role that shows empathy and a neutral curiosity to try to access the other's \textit{Jouissance}. It is also worth noting that the numerical values of $\mathbf{RI}$ are consistently lower than in all other cases. This reflects the fact that in daily dialogues people seldom adopt the neutral speech of an Analyst.}

\textbf{The top 5 emotions combinations in relation to the Analyst discourse are:}
\begin{itemize}
    \item \{approval, caring, curiosity\} $\mathbf{RI}$ \{A\} = 0.20;
    \item \{approval, caring, optimism\} $\mathbf{RI}$ \{A\} = 0.13;
    \item \{caring, curiosity\} $\mathbf{RI}$ \{A\} = 0.11;
    \item \{caring, curiosity, neutral\} $\mathbf{RI}$ \{A\} = 0.09;
    \item \{caring, neutral, realization\} $\mathbf{RI}$ \{A\} = 0.08.
\end{itemize}

\textbf{The only Hysteric discourse } is uniquely identified by a very large set of emotions combinations, 10 in total. The relation intensity is in the range:

\[
0.33 \leq \{e_i, e_j, e_k\} \mathbf{RI} \{H\} \leq 0.82,
\]

\noindent{and the emotions cover a wide spectrum with a prevalence of excitement. This finding is consistent with {\bf \$} in the position of \textit{Agent / Semblance} driven by {\bf a} in a role that looks forward to getting access to knowledge but expressing the power of his/her personal feelings. It is worth noting that the numerical values of $\mathbf{RI}$ are consistently higher than in all other cases. This reflects how frequent in the daily life people are in a position of being eagerly looking for information and understanding and also how this position is easier to be identified.}

\textbf{The top 5 emotions combinations in relation to the Hysteric discourse are:}
\begin{itemize}
    \item \{admiration, approval, excitement\} $\mathbf{RI}$ \{H\} = 0.82;
    \item \{approval, curiosity\} $\mathbf{RI}$ \{H\} = 0.75;
    \item \{disappointment, sadness\} $\mathbf{RI}$ \{H\} = 0.60;
    \item \{anxiety, curiosity, nervousness\} $\mathbf{RI}$ \{H\} = 0.50;
    \item \{anxiety, disapproval, nervousness\} $\mathbf{RI}$ \{H\} = 0.50. 
\end{itemize}

\textbf{The only University discourse} is uniquely identified by a set of 8 emotions combinations. The relation intensity is in the range:

\[
0.20 \leq \{e_i, e_j, e_k\} \mathbf{RI} \{U\} \leq 0.40,
\]

\noindent{and the emotions are characterized by neutrality, realization and judgement. This finding is consistent with {\bf S2} in the position of \textit{Agent / Semblance} driven by {\bf S1} in a role that acts as if restricted to deal with objective facts but at the same time looking for the fulfillment of institutional objectives ({\bf S2} is driven by {\bf S1}) and employing judgment to achieve such results. It is worth noting that numerical values of $\mathbf{RI}$ are quite similar to those obtained for the Master discourse.}

\textbf{The top 5 emotions combinations in relation to the University discourse are:}
\begin{itemize}
    \item \{neutral\} $\mathbf{RI}$ \{U\} = 0.78;
    \item \{confusion, neutral, realization\} $\mathbf{RI}$ \{U\} = 0.40;
    \item \{approval, realization\} $\mathbf{RI}$ \{U\} = 0.39;
    \item \{annoyance, neutral\} $\mathbf{RI}$ \{U\} = 0.27;
    \item \{joy, neutral\} $\mathbf{RI}$ \{U\} = 0.27.
\end{itemize}

\textbf{The only Capitalist discourse} is uniquely identified by only one combination of emotions, namely (approval, desire, joy). The relation intensity is equal to 0.20. It is likely that if the experiment is applied to a larger set of dialogues other combinations of emotions can be found. Nevertheless this set of emotions represents well the fact that {\bf \$} in the position of the \textit{Agent / Semblance} attempts to satisfy his internal needs not by addressing the \textit{other} but by directly accessing an ``artifact'' that represents his/her desire, would produce joy, would have his/her approval or would generate approvals to him/herself, for example, in terms of ``likes'' in the social media.

Many times a statement presents a mix of Lacanian discourses. For example, {\bf the combination \{Master, Hysteric\}} has appeared in 8 instances and relation intensity is in the range:

\[
0.26 \leq \{e_i, e_j, e_k\} \mathbf{RI} \{M, H\} \leq 0.89,
\]

\noindent{and the top 5 emotions combinations in relation to this mix of discourses are:}

\begin{itemize}
    \item \{annoyance, disappointment, disapproval\} $\mathbf{RI}$ \{M, H\} = 0.89;
    \item \{anxiety, realization\} $\mathbf{RI}$ \{M, H\} = 0.64;
    \item \{admiration, joy, surprise\} $\mathbf{RI}$ \{M, H\} = 0.50;
    \item \{annoyance, anxiety, embarrassment\} $\mathbf{RI}$ \{M, H\} = 0.50;
    \item \{caring, disappointment, embarrassment\} $\mathbf{RI}$ \{M, H\} = 0.50.
\end{itemize}

{\bf The combination \{Hysteric, University\}} has appeared in 5 instances and relation intensity is in the range:

\[
0.25 \leq \{e_i, e_j, e_k\} \mathbf{RI} \{H, U\} \leq 0.55,
\]

\noindent{and the emotions combinations in relation to this mix of discourses are:}

\begin{itemize}
    \item \{annoyance, disapproval, realization\} $\mathbf{RI}$ \{M, H\} = 0.55;
    \item \{confusion, disapproval, surprise\} $\mathbf{RI}$ \{M, H\} = 0.50;
    \item \{curiosity, desire, neutral\} $\mathbf{RI}$ \{M, H\} = 0.50;
    \item \{approval, disappointment\} $\mathbf{RI}$ \{M, H\} = 0.29;
    \item \{approval, joy, realization\} $\mathbf{RI}$ \{M, H\} = 0.25.
\end{itemize}

{\bf The combination \{Master, University\}} has appeared in 3 instances and relation intensity is in the range:

\[
0.13 \leq \{e_i, e_j, e_k\} \mathbf{RI} \{M, U\} \leq 0.71,
\]

\noindent{and the emotions combinations in relation to this mix of discourses are:}

\begin{itemize}
    \item \{approval, excitement, pride\} $\mathbf{RI}$ \{M, U\} = 0.71;
    \item \{annoyance, anxiety, disapproval\} $\mathbf{RI}$ \{M, U\} = 0.40;
    \item \{admiration, neutral, realization\} $\mathbf{RI}$ \{M, U\} = 0.13.
\end{itemize}

It is also worth noting that, for example, Anger -- a {\bf Basic Emotion} of the GoEmotions dataset -- has not appeared. At this stage, a possible explanation is that the set of dialogues employed in the experiment does not have this emotion but in a larger experiment it may appear.  Another possibility is that Anger may have been annotated either as Annoyance or Disapproval. Another case is Grief -- a {\bf Secondary Cognitive Emotion} of the GoEmotions dataset -- has not appeared as well. In this case, a possible explanation is that Grief has been annotated as Sadness.

It is left for the interested reader to go through the heat map and verify that the set of emotions associated to the combination of discourses are consistent with the Lacan's theory.

\section{Conclusions and Future Work}
\label{sec:discussion}
The main contribution of this research is a psychoanalytic one per se: the establishment of a systematic relation among emotions and Lacanian discourses. 

In Section \ref{sec:theoretical}, Theoretical Framework, the following topics have been discussed:

\begin{itemize}
    \item a review of how emotions appear in Freud’s works and their relation with the Unconscious;
    \item a review of how emotions appear in Lacan’s works;
    \item a review of the five Lacanian discourses;
    \item a description of the Lacanian Discourse Analysis (LDA).
    \item a review of the classification of emotions as proposed by different authors as, for example, Ekman and Cowen.
\end{itemize}

In Section \ref{sec:relatedwork}, Related Work, it is made a review of a few works that have some relationship with this work. To the best of our knowledge, there isn't any published work that tried to identify a relationship between emotions and the Lacanian Discourses. The discussed works are briefly reviewed not because they adopted similar approaches to this work and could be compared with but because of: (i) their background information that helps to understand the context in which this work has been developed; and (ii) their ideas that could be used to fuel future research.

In Section \ref{sec:adopted_methodology}, Adopted Methodology, a detailed description of all the steps involved in the methodology are described. Such steps comprehend:
\begin{itemize}
    \item Choice of the emotions set;
    \item Choice of working only with texts;
    \item Choice of using dialogues;
    \item Voting and dataset creation process;
    \item Probabilistic formulae to evaluate the relation between discourses and emotions.
\end{itemize}

In Section \ref{sec:results}, Results and Discussion, a comprehensive presentation of the:
\begin{itemize}
    \item Experimental procedure;
    \item Votes processing; and
    \item Findings and Results.
\end{itemize}

The main findings of this work can be summarized as follows:
\begin{itemize}
    \item An empirical evidence of the relationship between emotions and Lacanian Discourses has been established;
    \item Emotions have a strong differential power to identify unique Lacanian discourses present in texts;
    \item Emotions also have a strong differential power to identify mix of Lacanian discourses present in texts;
    \item The relationships that have been found can be explained psychoanalytically.
\end{itemize}

The findings of this work are limited due to the small set of dialogues used in the experimental procedure. However, the procedure has proved to be powerful and can be applied to a larger set to refine the results. We conjecture that new relationships will be found and those already found will be confirmed.

In terms of future work, it is worth noting that this method can be automatized to a great extent, since current computer-based methods (primarily employing AI systems and tools) can be used to effectively detect emotions in texts. This way, there is great potential for developing effective, real-world applications, based on the automated identification of emotions and corresponding discourses.

This work is just a first step of the Lacanian Discourse Discovery (LDD) methodology, that is presented for the first time here. In future, a structural approach will be developed to directly identify the signifiers {\bf S1, S2, \$, a} based on their unique properties without relying on emotions as an intermediate step.

\section*{Appendix}

\subsection{Pseudocodes}
\label{sec:pseudo}
\begin{algorithm}[htbp]
\caption{Pseudocode for Deriving Final Discourses for Sentences}
\label{alg:final_discourses}
\begin{algorithmic}[1]
\State \textbf{Input:} $V_1$, $V_2$, $V_3$, $N$, $D$
\State \textbf{Output:} $O$
\State Initialize the outcome set $O$ as an empty list
\State Create a frequency map $F$ for each discourse $d_i$ in $D$
\For {each $d_i$ in $V_1 \cup V_2 \cup V_3$}
    \If {$d_i$ in $V_1$} 
        \State Increment $F[d_i]$
    \EndIf
    \If {$d_i$ in $V_2$} 
        \State Increment $F[d_i]$
    \EndIf
    \If {$d_i$ in $V_3$} 
        \State Increment $F[d_i]$
    \EndIf
\EndFor
\For {each $d_i$ in $D$}
    \If {$F[d_i] == 3$} // Unanimous agreement
        \State Set Confidence = 'H'
        \If {all voters selected only $d_i$ without other discourses in $V_1$, $V_2$, $V_3$}
            \State Set Weight = 1.0
        \Else
            \State Decrement the weight by 0.2 for each discourse that appears with $d_i$ and appears only once in $V_1$, $V_2$, or $V_3$
        \EndIf
        \State Add $(d_i, \text{Confidence}, \text{Weight})$ to $O$
    \ElsIf {$F[d_i] == 2$} // Majority agreement
        \State Set Confidence = 'M'

        \If {no additional discourses selected by the voters or the additional discourses appear more than once}
            \State Set Weight = 1.0
        \Else
            \State Decrement the weight by 0.2 for each discourse that appears with $d_i$ and appears only once in $V_1$, $V_2$, or $V_3$
        \EndIf
        \State Add $(d_i, \text{Confidence}, \text{Weight})$ to $O$
    \ElsIf {$F[d_i] == 1$ and $d_i$ appears alone}
        \State Ignore $d_i$
    \Else
        \State Add $(d_i, \text{Confidence}, \text{Weight})$ to $O$
    \EndIf
\EndFor
\If {$N \geq 2$}
    \State Add $(\text{'none'}, L, 0)$ to $O$
\EndIf
\State Return $O$ as the final decision for sentence $s_i$
\end{algorithmic}
\end{algorithm}
\FloatBarrier

\begin{algorithm}[htbp]
\caption{Emotion Confidence Score Calculation}
\label{alg:emotion_confidence}
\begin{algorithmic}[1]
\State \textbf{Input:} 
\State \quad Voters = \{voter1, voter2, ..., voterN\}
\State \quad Emotions = \{em1, em2, ..., em10\}

\State \textbf{Output:} 
\State \quad ConfidenceScores: A list containing the normalized confidence score for each emotion in Emotions

\State \textbf{Algorithm:}
\State Initialize ConfidenceScores as an empty list with length equal to the number of emotions, initialized to zero

\State Create a mapping of confidence scores to numerical values:
\State \quad DN = 0
\State \quad PN = 1
\State \quad PY = 2
\State \quad DY = 3

\For {each voter in Voters}
   \For {each emotion in Emotions}
      \If {the voter has assigned a confidence score to the emotion}
         \State Retrieve the assigned score (DN, PN, PY, DY)
         \State Map the assigned score to its numerical value using the mapping
      \Else
         \State Implicitly assign DN to the emotion
         \State Map DN to its numerical value (0)
      \EndIf
   \EndFor
\EndFor

\State Create a table where each voter's confidence score for each emotion is replaced by its corresponding numerical value

\For {each emotion in Emotions}
   \State Initialize a variable Sum = 0
   \For {each voter}
      \State Add the voter's numerical value for the emotion to Sum
   \EndFor
   \State Normalize the confidence score by dividing Sum by the number of voters and then by 3 (value of DY)
   \State Store the normalized value in ConfidenceScores
\EndFor

\State \textbf{Return} ConfidenceScores as the final result for each emotion
\end{algorithmic}
\end{algorithm}
\FloatBarrier

\subsection{Screenshot of the platform}
\label{sec:sc_platform}
\begin{figure*}[!h]
    \centering
    \includegraphics[width=\textwidth]{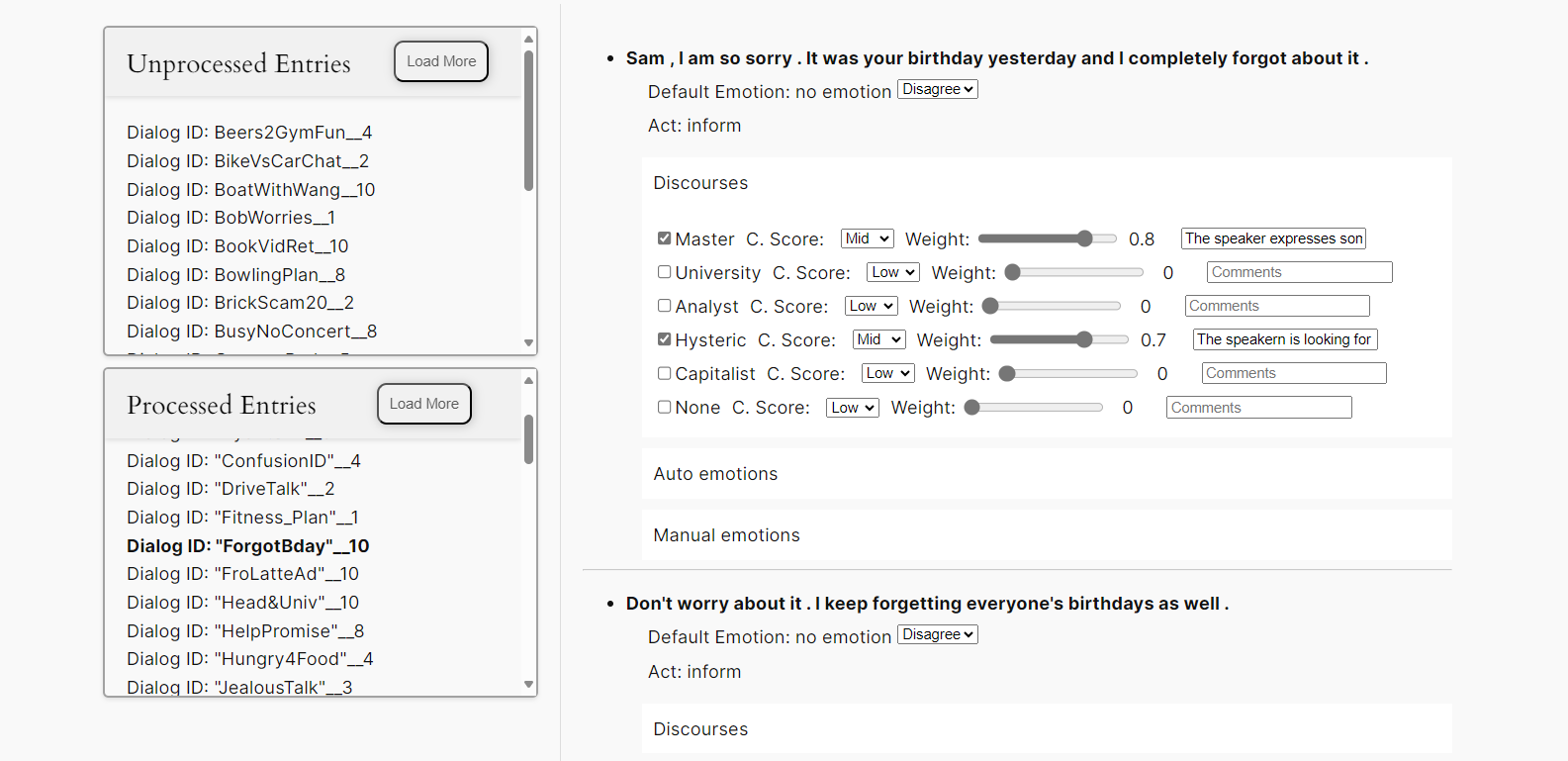}
    \caption{Screenshot of the platform}
    \label{fig:screenshot_platform}
\end{figure*}

A screenshot of the platform used for annotation is shown in Figure \ref{fig:screenshot_platform}
\vspace{0.5cm} 

\FloatBarrier

\section*{Conflict of Interest Statement}
The authors declare that the research was conducted in the absence of any commercial or financial relationships that could be construed as a potential conflict of interest.

\section*{Funding}
This research did not receive any funding.

\section*{Data Availability}
The datasets generated and analyzed during this study are available on GitHub at the following link: 
\url{https://github.com/PsyComp-Psychoanalytic-Driven-Computing/dailydialog_lacanian_discourses_emotions}.

\bibliographystyle{unsrt}  
\bibliography{references}  






\end{document}